\documentclass[a4paper,fleqn]{cas-sc}

\usepackage[square, comma, sort&compress, numbers]{natbib}

\usepackage[linesnumbered,ruled]{algorithm2e}
\usepackage{rotating}
\usepackage{url}
\def\tsc#1{\csdef{#1}{\textsc{\lowercase{#1}}\xspace}}
\tsc{WGM}
\tsc{QE}
\tsc{EP}
\tsc{PMS}
\tsc{BEC}
\tsc{DE}

\begin{document}
\let\WriteBookmarks\relax
\def\floatpagepagefraction{1}
\def\textpagefraction{.001}

% Short title
\shorttitle{Leveraging Large Language Model to Generate a Novel Metaheuristic Algorithm with CRISPE Framework}

% Short author
\shortauthors{Zhong R. et al.}

% Main title of the paper
\title [mode = title]{Leveraging Large Language Model to Generate a Novel Metaheuristic Algorithm with CRISPE Framework}            

\author[1]{Rui Zhong}[orcid=0000-0003-4605-5579]

\ead{rui.zhong.u5@elms.hokudai.ac.jp}
\credit{Conceptualization, Methodology, Investigation, Writing – original draft, Writing – review \& editing, and Funding acquisition.}
\affiliation[1]{organization={Graduate School of Information Science and Technology, Hokkaido University}, city={Sapporo}, country={Japan}}

\author[2]{Yuefeng Xu}[]
\ead{xyf20070623@gmail.com}
\credit{Methodology, Formal Analysis, and Writing – review \& editing.}
\affiliation[2]{organization={School of Engineering, University of Fukui}, city={Fukui}, country={Japan}}

\author[3]{Chao Zhang}[orcid=0000-0002-0845-9217]
\ead{zhang@u-fukui.ac.jp}
\credit{Investigation and Writing – review \& editing.}
\affiliation[3]{organization={Department of Engineering, University of Fukui}, city={Fukui}, country={Japan}}

\author[4]{Jun Yu}[orcid=0000-0001-5029-0294]
\cormark[1]
\ead{yujun@ie.niigata-u.ac.jp}
\credit{Conceptualization, Investigation, and Writing – review \& editing.}
\affiliation[4]{organization={Institute of Science and Technology, Niigata University}, city={Niigata}, country={Japan}}

% Corresponding author text
\cortext[cor1]{Corresponding author}

% Here goes the abstract
\begin{abstract}
In this paper, we borrow the large language model (LLM) ChatGPT-3.5 to automatically and quickly design a new metaheuristic algorithm (MA) with only a small amount of input. The novel animal-inspired MA named zoological search optimization (ZSO) draws inspiration from the collective behaviors of animals for solving continuous optimization problems. Specifically, the basic ZSO algorithm involves two search operators: the prey-predator interaction operator and the social flocking operator to balance exploration and exploitation well. Besides, the standard prompt engineering framework CRISPE (i.e., \textbf{C}apacity and \textbf{R}ole, \textbf{I}nsight, \textbf{S}tatement, \textbf{P}ersonality, and \textbf{E}xperiment) is responsible for the specific prompt design. Furthermore, we designed four variants of the ZSO algorithm with slight human-interacted adjustment. In numerical experiments, we comprehensively investigate the performance of ZSO-derived algorithms on CEC2014 benchmark functions, CEC2022 benchmark functions, and six engineering optimization problems. 20 popular and state-of-the-art MAs are employed as competitors. The experimental results and statistical analysis confirm the efficiency and effectiveness of ZSO-derived algorithms. At the end of this paper, we explore the prospects for the development of the metaheuristics community under the LLM era.
\end{abstract}

% Research highlights
\begin{highlights}
\item We introduce ChatGPT-3.5 into the metaheuristics community to design a novel animal-inspired metaheuristic algorithm.
\item The standard CRISPE framework is responsible for the specific prompt design.
\item Comprehensive numerical experiments are implemented to investigate the performance of ZSO-derived algorithms.
\item We discuss the potential for future development of the metaheuristics community under the LLM era.
\end{highlights}

% Keywords
% Each keyword is separated by \sep
\begin{keywords}
Large Language Model (LMM) \sep ChatGPT-3.5 \sep Metaheuristic algorithms (MAs) \sep zoological search optimization (ZSO) \sep CRISPE framework
\end{keywords}

\maketitle

\section{Introduction} \label{sec:1}
Generative artificial intelligence (AI) is an advanced AI technique that can generate text, images, music, or other media with similar structures to the training data \cite{Jovan:22, Longoni:22}. As the generative adversarial network (GAN) \cite{Goodfellow:14} and transformer \cite{Vaswani:17} were reported in 2014 and 2017, respectively, the rapid development of generative AI has attracted widespread attention from scholars. Especially in recent years, large model generative AI models have sprung up like mushrooms, such as Bing Chat \cite{Dao:23}, LLaMA \cite{Hugo:23, Peng:23}, DALL-E \cite{Aditya:21, Gary:22}, ChatGPT series \cite{Radford:18, Radford:19, Brown:20, OpenAI:23}, and Gemini \cite{Team:23}. These large language models (LLMs) have been widely applied and achieved great success in art, writing, education, product design, medicine, gaming, marketing, and other fields \cite{Yuan:22, Bhargavi:23, Pier:23, Arun:23}.

\begin{figure}[htbp]
    \centering
    \includegraphics[width=11cm]{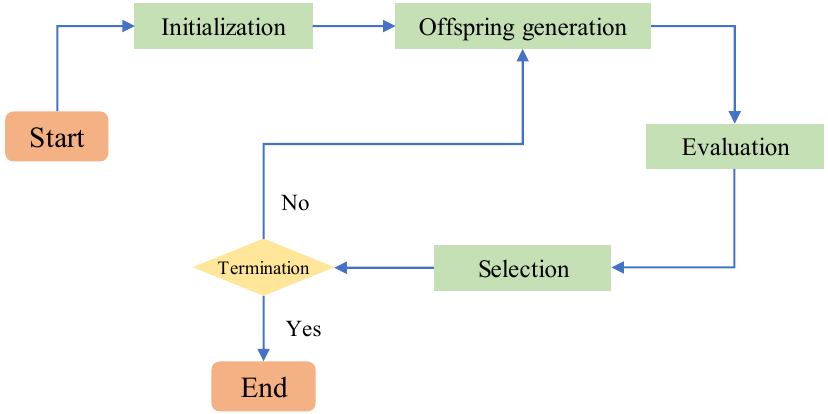}
    \caption{The general flowchart of EC techniques.}
    \label{fig:1.1}
\end{figure}

Evolutionary computation (EC), a sub-field of AI and soft computing, is a family of population-based stochastic optimization techniques inspired by natural phenomena or organism behaviors. A general flowchart of EC techniques is demonstrated in Figure \ref{fig:1.1}, encompassing five primary steps: (1). Initialization, which involves setting up the population and parameters, (2). Offspring generation, (3). Evaluation, (4). Selection, and (5). Iteration of step (2) to step (4) until the process reaches termination. Within these steps, the design of specialized search operators in step (2) stands as the core element of the EC technique, which can be realized by imitating the social behaviors of animals \cite{Zhong:23_3, Zitouni:22, Zhong:24} or by simulating the laws of science \cite{Rahmani:22, Daliri:22, Su:23}. Thanks to its scalability, easy implementation, robustness, and applicability, this approach has become one of the most favored optimization methodologies in both academic fields and real-world applications \cite{Zhong:23_1, Zhong:23_2}.

Additionally, the existence of the No Free Lunch Theorem (NFLT) \cite{Wolpert:97} suggests that the performance of any given metaheuristic algorithm (MA) cannot be guaranteed across all possible optimization problems. This uncertainty has sparked a surge in the development of the novel MAs. Therefore, the motivation of this research is to design a novel MA by integrating the LLM with the CRISPE framework. 

In this paper, we propose an LLM-assisted MA named the zoological search optimization (ZSO) algorithm. This innovative algorithm draws inspiration from the collective behavior and adaptation strategies observed in various animal species. It uniquely combines the concepts of social communication, dynamic collaboration, and individual adaptability to offer a unique metaheuristic approach for tackling black-box optimization problems. In the meantime, we address a critical limitation in the field of EC, as highlighted in \cite{Aranha:21}: the inadequate experimental validation often associated with newly proposed MAs. Thus, we implement a set of numerical experiments to investigate the performance of the proposed ZSO comprehensively. Moreover, we manually design four hyperparameter adaptation strategies and embed them into the ZSO. Experimental results and statistical analysis provide a practical verification of the high competitiveness of the proposed ZSO when compared to other expert-designed MAs. The specific contributions of this paper are outlined as follows:

\begin{itemize}
  \item Through the CRISPE framework and ChatGPT-3.5, we propose a novel zoological search optimization (ZSO) algorithm, specifically tailored for continuous optimization.
  \item We comprehensively investigate the performance of ZSO in CEC2014 benchmark functions, CEC2022 benchmark functions, and six engineering problems. 20 popular MAs are employed as competitors. 
  \item We introduce four hyperparameter adaptation variations within the ZSO framework.
  \item Both experimental and statistical results confirm the efficiency and robustness of the ZSO in various scenarios.
  \item This research also explores the future trajectory of the metaheuristics community in the era of the LLM.
\end{itemize}

The remainder of this paper is organized as follows: Section \ref{sec:2} involves the related works including the recent development of LLM-assisted EC techniques and an overview of the CRISPE framework. Section \ref{sec:3} covers a detailed introduction to the methodology in this study. Section \ref{sec:4} presents the numerical experiments conducted, along with the statistical results obtained from these optimization tests. Section \ref{sec:5} analyzes the performance of the proposed ZSO and provides some potential topics for EC promotion leveraging the LLM technique. Finally, Section \ref{sec:6} concludes the key findings and insights. 

\section{Related works} \label{sec:2}

\subsection{LLM-assisted metaheuristic algorithm} \label{sec:2.1}
The emergence of LLM has swiftly created a wave of interest in interdisciplinary research, making 'AI for science' a prominent term in contemporary discussions\cite{Liang:22, Igor:23}. However, the application of LLMs within the metaheuristics community remains relatively unexplored. A few notable exceptions include: Michal et al. \cite{Pluhacek:23} used the GPT-4 to generate a novel hybrid optimization approach based on particle swarm optimization (PSO), cuckoo search (CS), artificial bee colony (ABC), grey wolf optimizer (GWO), self-organizing migrating algorithm (SOMA), and whale optimization algorithm (WOA). This research primarily investigated the interactive process and challenges associated with using GPT-4 to fulfill a series of set tasks. Through the proper prompt engineering, Liu et al. \cite{Liu:23} discovered an efficient learning-based search operator by the LLM and proposed a new variant of decomposition-based multiobjective evolutionary algorithms named MOEA/D-LO, which demonstrates competitiveness with the expert-designed MOEAs. In another study, Liu et al. \cite{Fei:23} also proposed a methodology of algorithm evolution using LLM (AEL). Distinct from earlier methods, AEL is preferred to be determined as a paradigm rather than a problem-specific approach. It does not require model training and focuses on algorithm-level evolution, showing promising results in the discrete optimization domain such as the traveling salesman problem (TSP). Similarly, Liu et al. \cite{Sheng:23} proposed an LLM-driven evolutionary algorithm (LMEA) for solving TSPs. In each iteration, LMEA instructs the LLM to select parent solutions from the population and performs search operators to generate offspring solutions. Then, these offspring solutions are evaluated and selected for the next generation. Notably, the proposed LMEA has shown competitive performance in TSP instances with up to 20 cities. 

\subsection{CRISPE framework} \label{sec:2.2}
\begin{figure}[htbp]
    \centering
    \includegraphics[width=13cm]{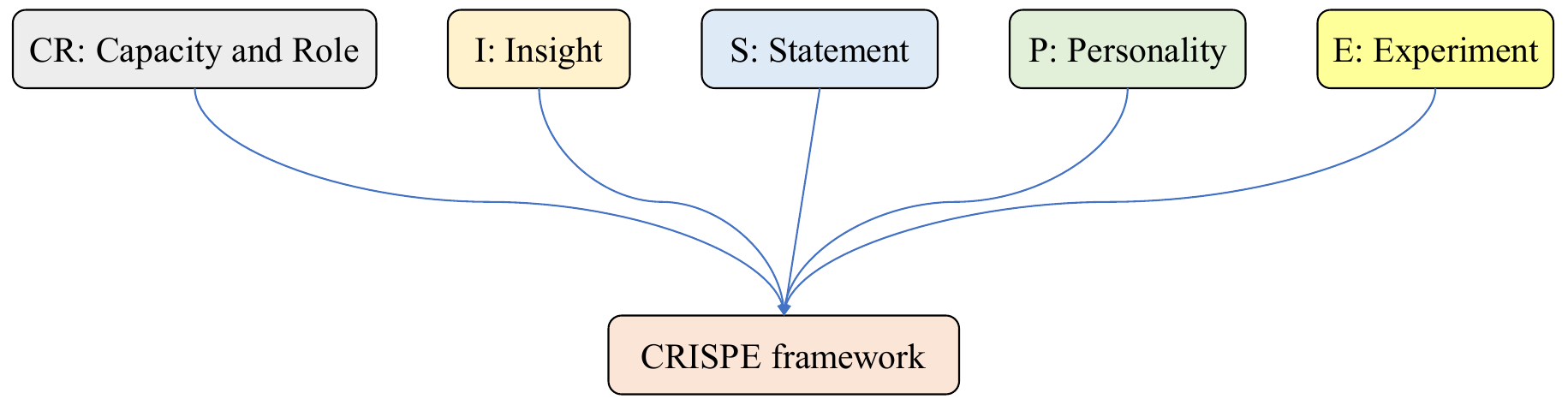}
    \caption{Components in the CRISPE framework.}
    \label{fig:2.2.1}
\end{figure}

As the LLM becomes fashionable in many research fields, the design of prompts to achieve desired outcomes also attracts the attention of scholars, which promotes the development of the terminology of "prompt engineering". Some representative methods are zero/one/few-shot prompting, iterative prompting, chain-of-thought prompting, and hybrid prompting. In this study, the hybrid approach CRISPE\cite{Wang:23, Shieh:23} is employed as the framework which contains five components and is demonstrated in Figure \ref{fig:2.2.1}.

\begin{itemize}
  \item CR: Define the capacity and role.
  \item I: Provide necessary background or insight.
  \item S: State the core of the prompt.
  \item P: Define the fashion of the LLM's response.
  \item E: Ask for multiple responses.
\end{itemize}

\section{Methodology in this study} \label{sec:3}
\begin{figure}[htbp]
    \centering
    \includegraphics[width=13cm]{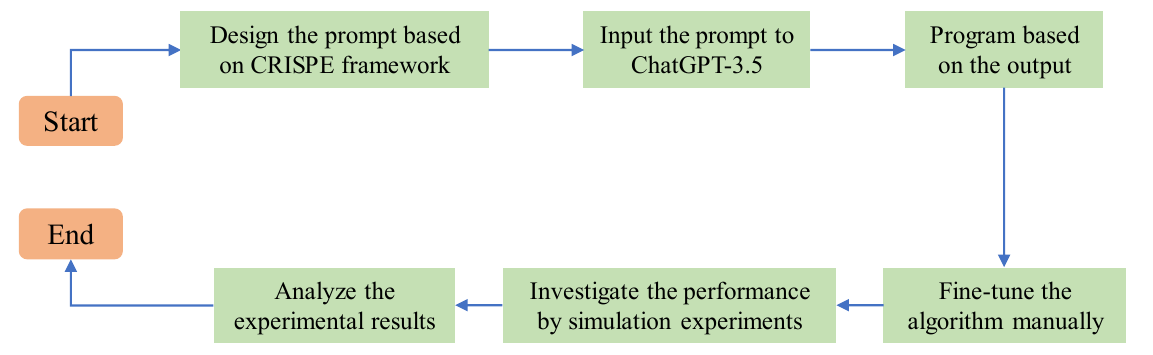}
    \caption{The flowchart of LLM-assisted metaheuristic design.}
    \label{fig:3.1}
\end{figure}

In this section, we introduce the process flow of designing an MA assisted by LLMs, as depicted in Figure \ref{fig:3.1}. The procedure commences with prompt engineering, where we carefully craft a suitable prompt guided by the CRISPE framework. Then, we input the prompt to ChatGPT-3.5. The output received from ChatGPT-3.5 forms the basis for programming the MA, followed by a manual fine-tuning of its hyperparameters. In the following context, we sequentially introduce the CRISPE-based prompt design and the generated MA: ZSO.

\subsection{CRISPE-based prompt design} \label{sec:3.1}
\begin{figure}[htbp]
    \centering
    \includegraphics[width=10cm]{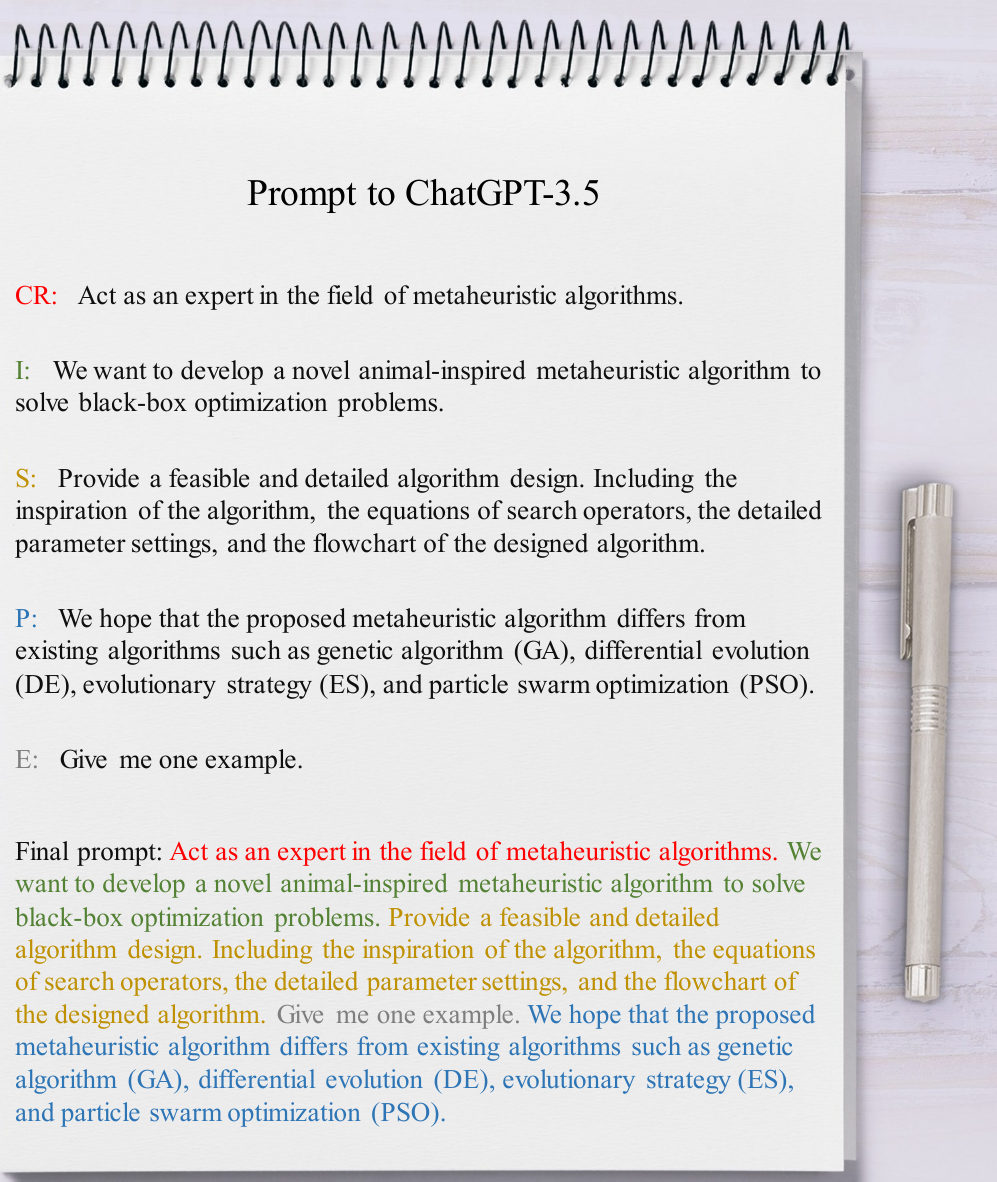}
    \caption{The CRISPE-based prompt design.}
    \label{fig:3.1.1}
\end{figure}

Figure \ref{fig:3.1.1} demonstrates the designed prompt based on the CRISPE framework for use with ChatGPT-3.5. This prompt is designed to engage ChatGPT-3.5 as though it were an expert in the field of MAs. In the insight component, we convey to ChatGPT-3.5 our intention to develop a novel animal-inspired MA to solve black-box optimization problems. In the statement component, ChatGPT-3.5 is required to provide a feasible and detailed algorithm design, including the inspiration of the algorithm, the equations of search operators, the detailed parameter settings, and the flowchart of the designed algorithm. In the personality part, the focus here is on innovation and distinction. We express our desire for the proposed MA to be distinct from existing algorithms such as genetic algorithm (GA), differential evolution (DE), evolutionary strategy (ES), and particle swarm optimization (PSO). Finally, only one example is requested by ChatGPT-3.5. In summary, the complete prompt can be found in Figure \ref{fig:3.1.1} clearly.

\subsection{Zoological search optimization (ZSO)} \label{sec:3.2}
\begin{figure}[htbp]
    \centering
    \includegraphics[width=11cm]{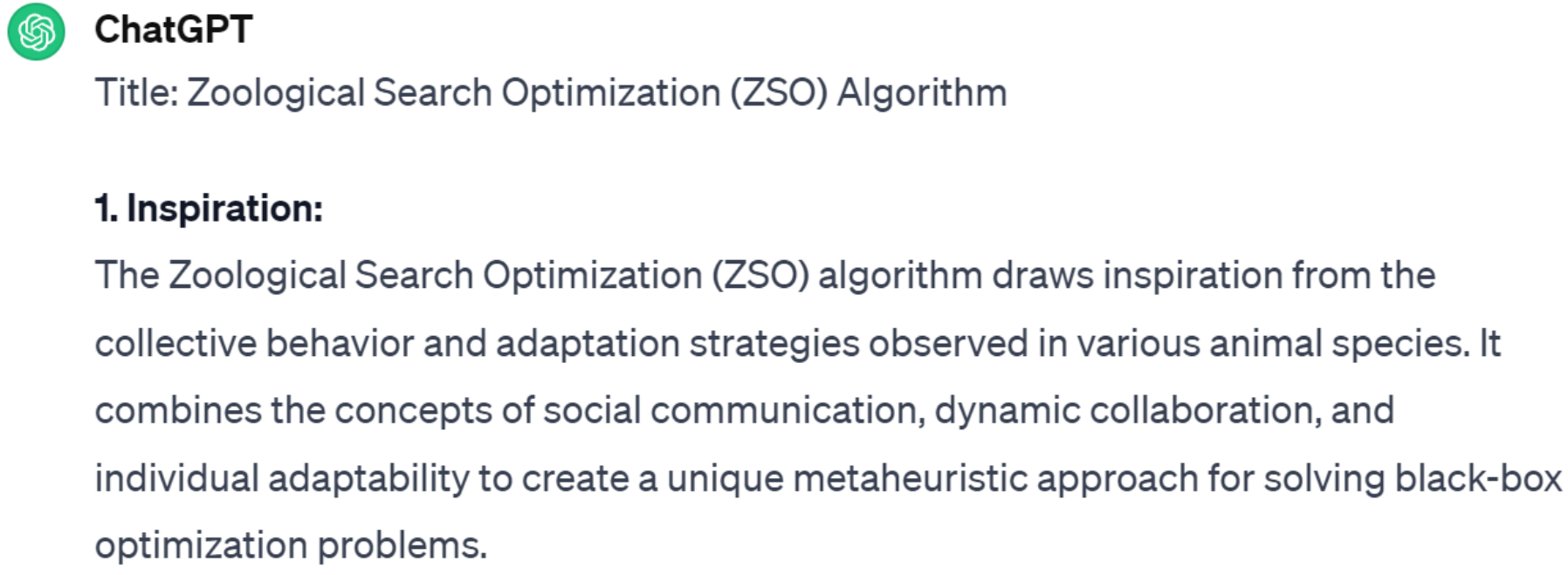}
    \caption{The inspiration of ZSO.}
    \label{fig:3.2.1}
\end{figure}

\begin{figure}[htbp]
    \centering
    \includegraphics[width=7cm]{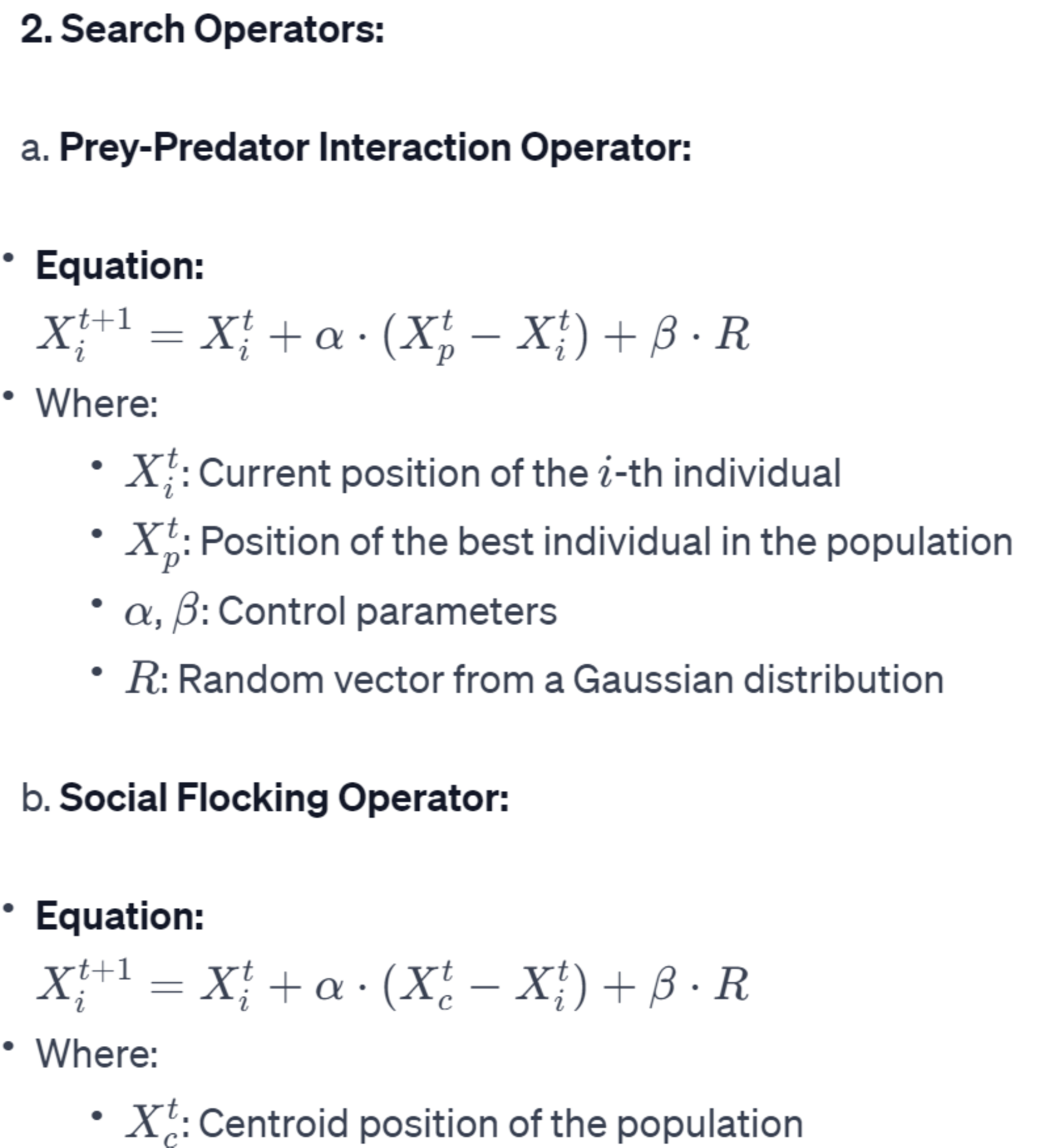}
    \caption{The designed search operators of ZSO.}
    \label{fig:3.2.2}
\end{figure}

\begin{figure}[htbp]
    \centering
    \includegraphics[width=10cm]{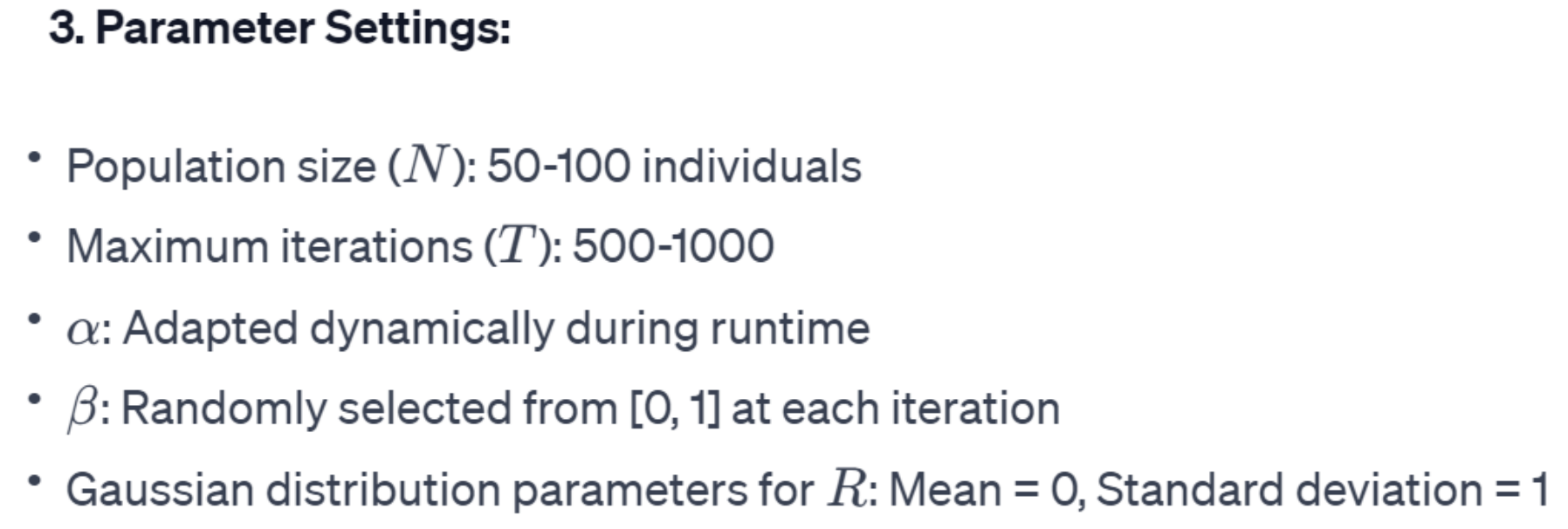}
    \caption{The parameter settings of ZSO.}
    \label{fig:3.2.3}
\end{figure}

\begin{figure}[htbp]
    \centering
    \includegraphics[width=10cm]{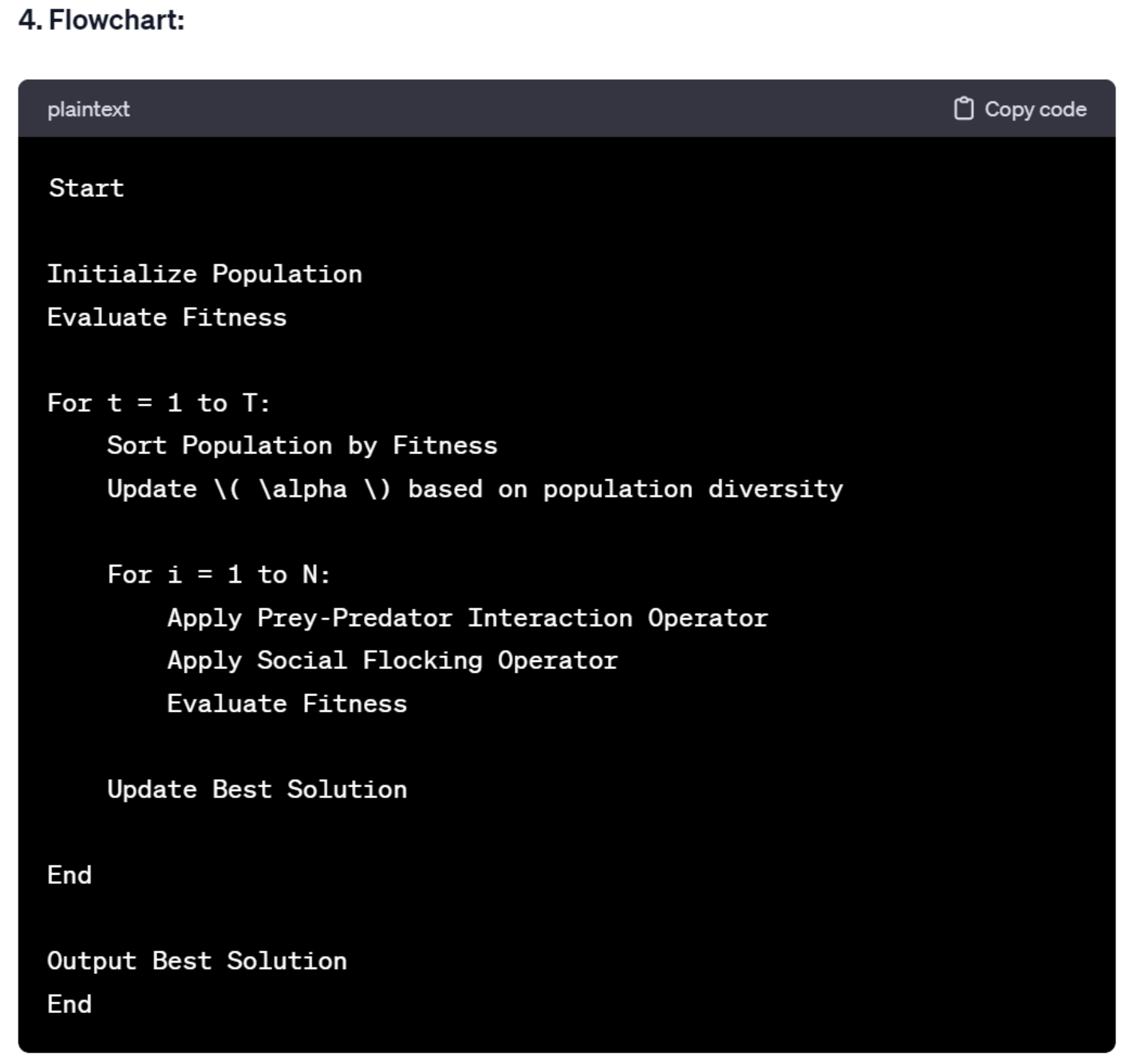}
    \caption{The designed search operators of ZSO.}
    \label{fig:3.2.4}
\end{figure}

\begin{figure}[htbp]
    \centering
    \includegraphics[width=10cm]{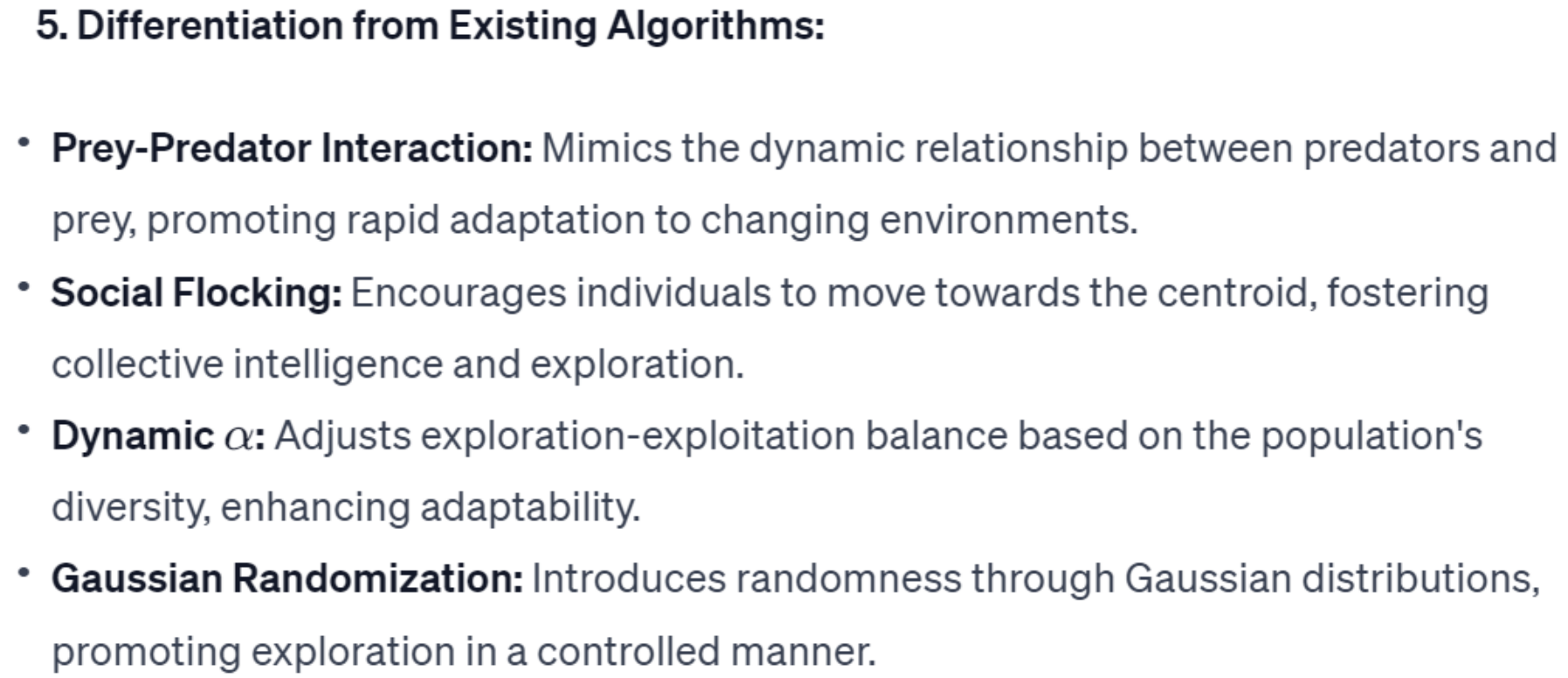}
    \caption{The difference between ZSO and existing MAs.}
    \label{fig:3.2.5}
\end{figure}

The results generated by ChatGPT-3.5 elucidate the ZSO methodology, which encompasses five components: the inspiration, the search operators, the parameter settings, the flowchart of the pseudocode, and the difference between the existing MAs. These elements are systematically delineated in Figures \ref{fig:3.2.1}, \ref{fig:3.2.2}, \ref{fig:3.2.3}, \ref{fig:3.2.4}, and \ref{fig:3.2.5}, corresponding to each component respectively.

Focusing on the search operators presented in Figure \ref{fig:3.2.4}, since two control hyperparameters $\alpha$ and $\beta$ are not recommended with specific values by ChatGPT-3.5, we have developed four variants: ZSO with constant hyperparameters (ZSO\_Cons), ZSO with linear decreasing hyperparameters (ZSO\_Linear), ZSO with uniform random hyperparameters (ZSO\_Uniform), and ZSO with Gaussian random hyperparameters (ZSO\_Gauss). Detailed descriptions of the adaptive strategies for each variant are set to be presented in Section \ref{sec:4.1.2}. Moreover, we have incorporated an additional evaluation process subsequent to both the prey-predator interaction operator and the social flocking operator within the ZSO framework. This modification allows ZSO to update the current population more frequently, but under the constraints of fixed maximum fitness evaluations (FEs), the maximum iterations will be reduced by half.

\section{Numerical experiments} \label{sec:4}
This section provides a comprehensive overview of the detailed experiment settings and the corresponding results. Section \ref{sec:4.1} is dedicated to outlining the specific details of the experiment settings: experimental environments, benchmark functions, and compared methods with their parameters. Section \ref{sec:4.2} presents the experimental results and statistical analysis. 

\subsection{Experiment settings} \label{sec:4.1}
\subsubsection{Experimental environments and implementation} \label{sec:4.1.4}
All MAs are implemented using Python 3.11 and executed on the Lenovo Legion R9000P, operating under Windows 11. The system's hardware configuration includes an AMD Ryzen 7 5800H processor with Radeon Graphics clocked at 3.20 GHz and 16GB RAM. Notably, except for ZSO and its variants, all competitor MAs were implemented using the MEALPY library \cite{Nguyen:23}. The CEC2014 \cite{Liang:14} and CEC2022 \cite{Abhishek:22} benchmark functions used for the evaluation are sourced from the OpFuNu library \cite{Thieu:20}. Additionally, six engineering optimization problems\cite{Ezugwu:22} are obtained from the ENOPPY library \cite{Thieu:23}.

\subsubsection{Benchmark functions} \label{sec:4.1.1}
To evaluate our proposed ZSO and its variants comprehensively and fairly, we have structured our optimization experiments to encompass four distinct types of optimization problems: 

(1). 30-D and 50-D CEC2014 benchmark functions in Table \ref{tbl:4.1.1.1}, which investigates the performance of MAs in median-scale tasks.
\begin{table}[htbp]
	\scriptsize
	\centering
	\renewcommand\arraystretch{1.3}
	\caption{Summary of the CEC2014 suite: Uni.=Unimodal function, Multi.=Multimodal function, Hybrid.=Hybrid function, Comp.=Composition function}
	\label{tbl:4.1.1.1}
	\begin{tabular}{cccc}
		\toprule
		Func. & Description & Feature & Optimum  \\
		\midrule
            $f_1$ & Rotated High Conditioned Elliptic Function & \multirow{3}{*}{Uni.} & 100 \\
            $f_2$ & Rotated Bent Cigar Function & ~ & 200 \\
            $f_3$ & Rotated Discus Function & ~ & 300 \\
        \midrule
            $f_4$ & Shifted and Rotated Rosenbrock’s Function & \multirow{13}{*}{Multi.} & 400 \\
            $f_5$ & Shifted and Rotated Ackley’s Function & ~ & 500 \\
            $f_6$ & Shifted and Rotated Weierstrass Function &   & 600 \\
            $f_7$ & Shifted and Rotated Griewank’s Function & ~ & 700 \\
            $f_8$ & Shifted Rastrigin’s Function & ~ & 800 \\
            $f_9$ & Shifted and Rotated Rastrigin’s Function & ~ & 900 \\
            $f_{10}$ & Shifted Schwefel’s Function & ~ & 1000 \\
            $f_{11}$ & Shifted and Rotated Schwefel’s Function & ~ & 1100 \\
            $f_{12}$ & Shifted and Rotated Katsuura Function & ~ & 1200 \\
            $f_{13}$ & Shifted and Rotated HappyCat Function & ~ & 1300 \\
            $f_{14}$ & Shifted and Rotated HGBat Function & ~ & 1400 \\
            $f_{15}$ & Shifted and Rotated Expanded Griewank’s plus Rosenbrock’s Function  & ~ & 1500 \\
            $f_{16}$ & Shifted and Rotated Expanded Scaffer’s F6 Function & ~ & 1600 \\
        \midrule
            $f_{17}$ & Hybrid Function 1 ($N$=3) & \multirow{6}{*}{Hybrid.} & 1700 \\
            $f_{18}$ & Hybrid Function 2 ($N$=3) & ~ & 1800 \\
            $f_{19}$ & Hybrid Function 3 ($N$=4) & ~ & 1900 \\
            $f_{20}$ & Hybrid Function 4 ($N$=4) & ~ & 2000 \\
            $f_{21}$ & Hybrid Function 5 ($N$=5) & ~ & 2100 \\
            $f_{22}$ & Hybrid Function 6 ($N$=5) & ~ & 2200 \\
        \midrule
            $f_{23}$ & Composition Function 1 (n=5) & \multirow{8}{*}{Comp.} & 2300 \\
            $f_{24}$ & Composition Function 2 (n=3) & ~ & 2400 \\
            $f_{25}$ & Composition Function 3 (n=3) & ~ & 2500 \\
            $f_{26}$ & Composition Function 4 (n=5) & ~ & 2600 \\
            $f_{27}$ & Composition Function 5 (n=5) & ~ & 2700 \\
            $f_{28}$ & Composition Function 6 (n=5) & ~ & 2800 \\
            $f_{29}$ & Composition Function 7 (n=3) & ~ & 2900 \\
            $f_{30}$ & Composition Function 8 (n=3) & ~ & 3000 \\
        \midrule
            \multicolumn{4}{c}{Search range: [-100, 100]$^D$} \\
		\bottomrule
	\end{tabular}
\end{table}

(2). 10-D and 20-D CEC2022 benchmark functions in Table \ref{tbl:4.1.1.2}, which allows the performance evaluation in low-dimensional problems.
\begin{table}[htbp]
	\scriptsize
	\centering
	\renewcommand\arraystretch{1.3}
	\caption{Summary of the CEC2022 benchmark functions: Uni.=Unimodal function, Basic.=Basic function, Hybrid.=Hybrid function, Comp.=Composition function}
	\label{tbl:4.1.1.2}
	\begin{tabular}{cccc}
		\toprule
		Func. & Description & Feature & Optimum  \\
		\midrule
            $f_1$ & Shifted and full Rotated Zakharov & Uni. & 300 \\
            \midrule
            $f_2$ & Shifted and full Rotated Rosenbrock & \multirow{4}{*}{Basic.} & 400 \\
            $f_3$ & Shifted and full Rotated Expanded Schaffer $f_6$ & ~ & 600 \\
            $f_4$ & Shifted and full Rotated Non-Continuous Rastrigin & ~ & 800 \\
            $f_5$ & Shifted and full Rotated Levy & ~ & 900 \\
            \midrule
            $f_6$ & Hybrid function 1 (N = 3) & \multirow{3}{*}{Hybrid.} & 1800 \\
            $f_7$ & Hybrid function 2 (N = 6) & ~ & 2000 \\
            $f_8$ & Hybrid function 3 (N = 5) & ~ & 2200 \\
            \midrule
            $f_9$ & Composition function 1 (N = 5) & \multirow{4}{*}{Comp.}  & 2300 \\
            $f_{10}$ & Composition function 2 (N = 4) & ~ & 2400 \\
            $f_{11}$ & Composition function 3 (N = 5) & ~ & 2600 \\
            $f_{12}$ & Composition function 3 (N = 6) & ~ & 2700 \\
        \midrule
            \multicolumn{4}{c}{Search range: [-100, 100]$^D$} \\
		\bottomrule
	\end{tabular}
\end{table}

(3). Six real-world engineering optimization problems in Table \ref{tbl:4.1.1.3}, which evaluates the performance of optimizers in real-world scenarios. The specific visualization and equation can be found in \cite{Hadi:21}.

\begin{table}[htbp]
	\scriptsize
	\centering
	\renewcommand\arraystretch{1.2}
	\caption{Summary of six engineering optimization problems.}
	\label{tbl:4.1.1.3}
	\begin{tabular}{cccc}
		\toprule
		Name & Abbr. & Dim. & \# of constraints \\
		\midrule
            Cantilever Beam Design & CBD & 5 & 1 \\
            Corrugated Bulkhead Design & CBHD & 4 & 6 \\
            Gear Train Design & GTD & 4 & 0 \\
            I Beam Design & IBD & 4 & 2 \\
            Reinforced Concrete Beam & RCB & 3 & 2 \\
            Speed Reducer Design & SRD & 7 & 11 \\
		\bottomrule
	\end{tabular}
\end{table}

\subsubsection{Compared methods and parameters} \label{sec:4.1.2}
To comprehensively evaluate the performance of ZSO and its variants, 20 competitor MAs are employed including the genetic algorithm (GA) \cite{John:92}, particle swarm optimization (PSO) \cite{Kennedy:95}, differential evolution (DE) \cite{Storn:97}, evolution strategy with covariance matrix adaptation (CMA-ES) \cite{Hansen:03}, grey wolf optimizer (GWO)\cite{Seyedali:14}, moth flame optimization (MFO) \cite{Mirjalili:15}, Jaya optimizer \cite{Venkata:16}, sine cosine algorithm (SCA) \cite{Seyedali:16}, whale optimization algorithm (WOA) \cite{Andrew:16}, Harris hawks optimization (HHO) \cite{Ali:19}, Aquila optimizer (AO) \cite{Laith:21}, Archimedes optimization algorithm (ArchOA) \cite{Hashim:21}, arithmetic optimization algorithm (AOA) \cite{Abualigah:21}, serval optimization algorithm (SOA) \cite{Dehghani:22}, golden jackal optimization (GJO) \cite{Nitish:22}, fox optimizer (FOX) \cite{Mohammed:22}, energy valley optimizer (EVO) \cite{Azizi:23}, sea-horse optimizer (SHO) \cite{Zhao:22}, Chernobyl disaster optimizer (CDO) \cite{Shehadeh:23}, and coati optimization algorithm (COA) \cite{Mohammad:23}. The population size of all algorithms is set to 100. In terms of the maximum number of fitness evaluations (FEs), we have set specific criteria based on the type of optimization problem. For the CEC2014 and CEC2022 benchmark functions, the maximum FEs is 1000 $\cdot$ D. For the engineering optimization problems, we have set the maximum FEs at 20,000. To alleviate the randomness in the optimization, each competitor algorithm undergoes 30 independent trial runs. The detailed parameter settings for the compared methods are listed in Table \ref{tbl:4.1.2.1}. All parameters are consistent with the suggested settings in corresponding papers.
\begin{table}[htbp]
	\scriptsize
	\centering
	\renewcommand\arraystretch{1.4}
	\caption{The parameters of all compared optimization methods}
	\label{tbl:4.1.2.1}
	\resizebox{0.9\columnwidth}{!}{
		\begin{tabular}{ccc}
            \toprule
                Alg. & Parameters and Value \\
            \midrule
                GA & Crossover probability $pc$: 0.95; Mutation probability $pm$: 0.025; Selection scheme: tournament \\
                PSO & Inertia factor $w$: 1; Acceleration coefficients $c_1$ and $c_2$: 2.05; Max. and min. speed: 2 and -2 \\
                DE & Scaling factor $F$: 0.8; Crossover rate $Cr$: 0.9; Strategy: DE/cur-to-best/1/bin \\ 
                CMA-ES & Hyperparameter-free \\ 
                GWO & Hyperparameter-free \\ 
                MFO & Hyperparameter-free \\ 
                Jaya & Hyperparameter-free \\ 
                SCA & Constant $A$: 2 \\
                WOA & Constant $b$: 1 \\
                HHO & Hyperparameter-free \\ 
                AO & Hyperparameter-free \\ 
                ArchOA & Constants $c_1$, $c_2$, $c_3$, and $c_4$: 2, 5, 2, and 2; Max. and min. accelerator: 0.9 and 0.1 \\
                AOA & Constants $\alpha$ and $\mu$: 5 and 0.5; Max. and min. accelerator: 0.9 and 0.2 \\
                SOA & Hyperparameter-free \\ 
                GJO & Hyperparameter-free \\ 
                FOX & Jumping probabilities $c_1$ and $c_2$: 0.18 and 0.82 \\
                EVO & Hyperparameter-free \\ 
                SHO & Hyperparameter-free \\ 
                CDO & Constants $S_{\alpha}$, $S_{\beta}$, and $S_{\gamma}$: 16,000, 270,000, and 300,000 \\
                COA & Hyperparameter-free \\ 
		  \bottomrule
		\end{tabular}
  }
\end{table}

Additionally, the hyperparameters $\alpha$ and $\beta$ in the ZSO are manually designed with four strategies, which are listed in the following context:

\textbf{Constant}: $\alpha=1$ and $\beta=1$.

\textbf{Linear}: $\alpha=2 \cdot (1 - t/T_{max})$ and $\beta=2 \cdot (1 - t/T_{max})$, where $t$ is the current iteration and $T_{max}$ denotes the maximum iteration.

\textbf{Uniform}: $\alpha=\textbf{rand()}$ and $\beta=\textbf{rand()}$, where \textbf{rand()} generates a random value follows the distribution of $U(0.5, 1.5)$.

\textbf{Gaussian}: $\alpha=\textbf{randn()}$ and $\beta=\textbf{randn()}$, where \textbf{randn()} generates a random value follows the distribution of $N(1, 0.5)$.

Since the engineering problem contains constraints, the original MAs including ZSO cannot solve constrained optimization problems. Therefore, we equip all MAs with the static penalty function \cite{Carlos:02}, which is formulated in Eq. (\ref{eq:4.1.2.1})
\begin{equation}
    \label{eq:4.1.2.1}
    \begin{aligned}
        F(R_{i}) = f(R_{i}) + w\cdot \sum^m_{i=1}(\max(0, g_i(R_{i})))
    \end{aligned}
\end{equation}
$F(\cdot)$ is the fitness function, $f(\cdot)$ and $g_i(\cdot)$ are the objective function and constraint function, respectively. $w$ is a constant set to $10e7$ by default.

\subsection{Experimental results} \label{sec:4.2}
This section provides the experimental results and performs statistical analysis. We first present convergence curves of all MAs applied to the optimization problems, then, the statistical results are summarized by using the U-test and the Holm multiple comparison test. The marks $+$, $\approx$, and $-$ are used to indicate that our proposed ZSO\_Gauss is significantly better, has no significant difference, or is significantly worse compared to each of the competitor algorithms. Additionally, the average rank for each algorithm is determined.

\subsubsection{Results on CEC2014 benchmark functions} \label{sec:4.2.1}
\begin{figure}[!ht]
    \centering
    \includegraphics[width=16cm]{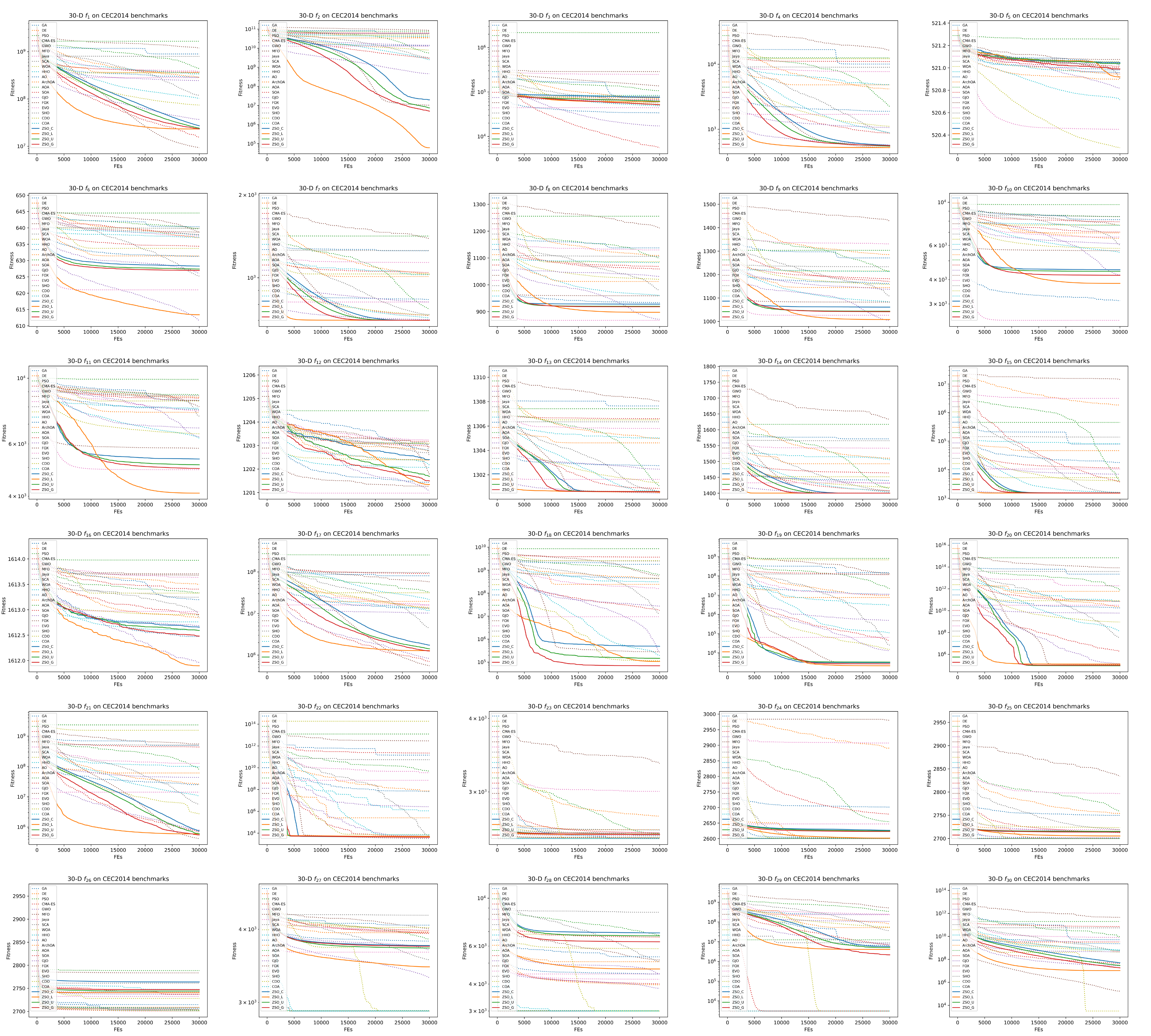}
    \caption{Convergence curves of competitor MAs on 30-D CEC2014 benchmark functions.}
    \label{fig:4.2.1.1}
\end{figure}

\begin{figure}[!ht]
    \centering
    \includegraphics[width=16cm]{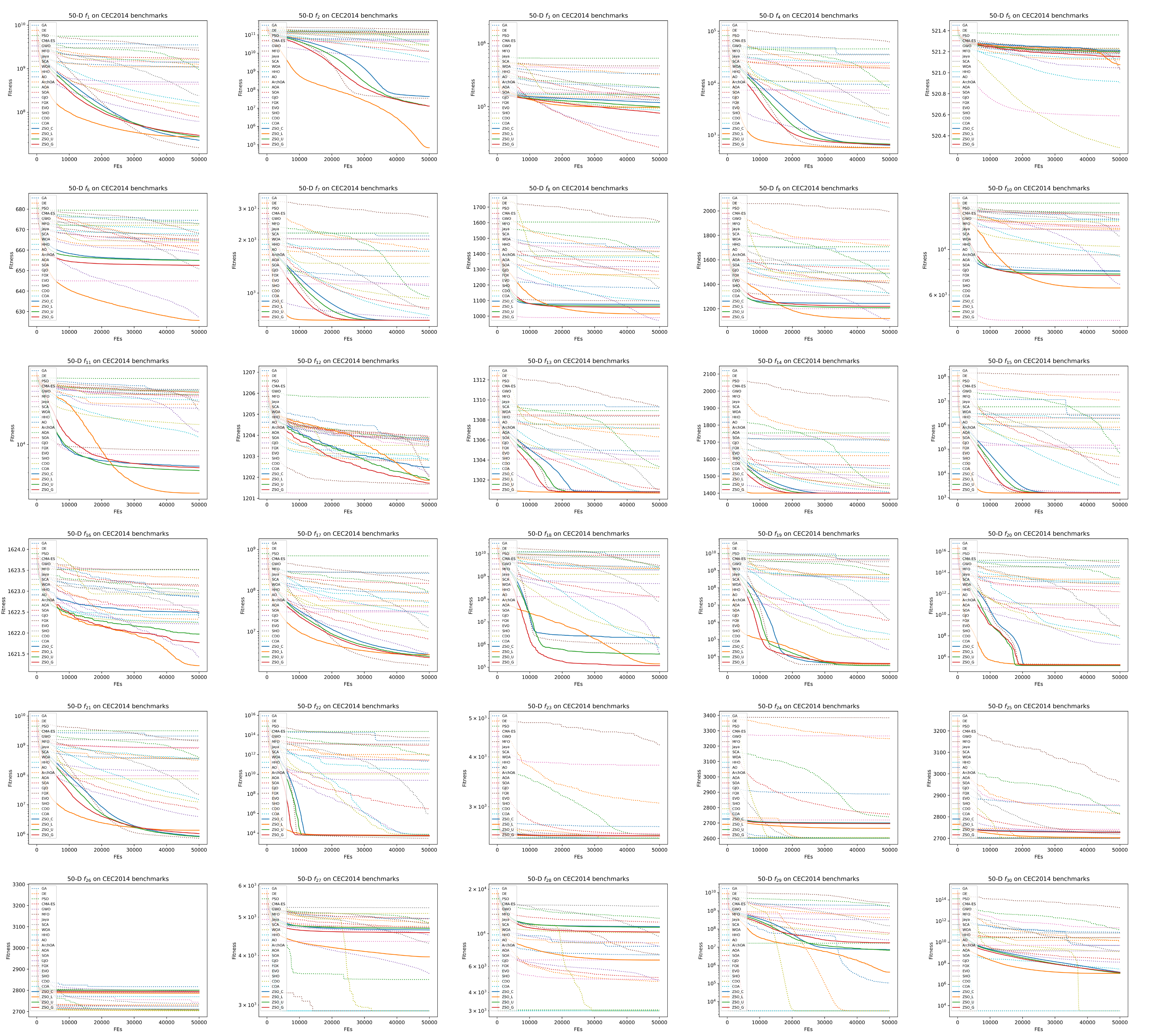}
    \caption{Convergence curves of competitor MAs on 50-D CEC2014 benchmark functions.}
    \label{fig:4.2.1.2}
\end{figure}

Figures \ref{fig:4.2.1.1} and \ref{fig:4.2.1.2} visualize the convergence curves of competitor algorithms on CEC2014 benchmark functions, Tables \ref{tbl:4.2.1.1} and \ref{tbl:4.2.1.2} summarize the corrected p-value and significance level between ZSO\_Gauss with other competitor algorithms.
\begin{sidewaystable}[htbp]
	\scriptsize
	\centering
	\renewcommand\arraystretch{1.5}
	\caption{Experimental and statistical results between ZSO\_Gauss and other competitor algorithms on 30-D CEC2014 benchmark functions. The mean value of 30 trial runs, the corrected p-value, and the significance level are provided. ZSO\_C, ZSO\_L, ZSO\_U, and ZSO\_G are abbreviations of ZSO\_Cons, ZSO\_Linear, ZSO\_Uniform, and ZSO\_Gauss.}
	\label{tbl:4.2.1.1}
	\resizebox{\columnwidth}{!}{
		\begin{tabular}{ccccccccccccccccccccccccc}
			\toprule
			\multirow{2}{*}{Func.} & GA & DE & PSO & CMA-ES & GWO & MFO & Jaya & SCA & WOA & HHO & AO & ArchOA \\
                ~ & AOA & SOA & GJO & FOX & EVO & SHO & CDO & COA & ZSO\_C & ZSO\_L & ZSO\_U & ZSO\_G \\
			\midrule
			\multirow{2}{*}{$f_1$} & 2.892e+08 $+$ & 2.867e+08 $+$ & 4.322e+08 $+$ & 1.548e+07 $-$ & 3.707e+07 $+$ & 1.199e+09 $+$ & 3.851e+08 $+$ & 1.204e+08 $+$ & 7.444e+07 $+$ & 1.031e+08 $+$ & 8.981e+08 $+$ & 3.664e+08 $+$ \\
                ~ & 1.671e+09 $+$ & 3.687e+08 $+$ & 2.460e+08 $+$ & \textbf{9.322e+06} $-$ & 2.848e+08 $+$ & 7.979e+08 $+$ & 3.574e+08 $+$ & 3.398e+08 $+$ & 2.710e+07 $\approx$ & 2.318e+07 $\approx$ & 2.410e+07 $\approx$ & 2.320e+07 \\
			\multirow{2}{*}{$f_2$} & 1.352e+10 $+$ & 3.696e+10 $+$ & 5.567e+09 $+$ & 2.908e+09 $+$ & 4.645e+08 $+$ & 8.531e+10 $+$ & 5.971e+10 $+$ & 3.239e+09 $+$ & 7.998e+09 $+$ & 2.513e+09 $+$ & 6.180e+10 $+$ & 3.467e+10 $+$ \\
                ~ & 7.712e+10 $+$ & 6.264e+10 $+$ & 1.439e+10 $+$ & 1.035e+07 $+$ & 1.273e+10 $+$ & 5.673e+10 $+$ & 4.871e+10 $+$ & 4.168e+10 $+$ & 1.972e+07 $+$ & \textbf{6.111e+04} $-$ & 7.472e+06 $+$ & 5.045e+06 \\
			\multirow{2}{*}{$f_3$} & 3.400e+04 $-$ & 7.111e+04 $+$ & 1.071e+05 $+$ & \textbf{5.589e+03} $-$ & 1.728e+04 $-$ & 2.890e+05 $+$ & 2.487e+05 $+$ & 5.646e+04 $\approx$ & 5.288e+04 $\approx$ & 4.912e+04 $\approx$ & 7.515e+04 $+$ & 7.462e+04 $+$ \\
                ~ & 2.172e+06 $+$ & 6.728e+04 $+$ & 6.529e+04 $+$ & 1.371e+05 $+$ & 4.462e+04 $-$ & 7.833e+04 $+$ & 8.297e+04 $+$ & 8.297e+04 $+$ & 7.622e+04 $+$ & 6.031e+04 $+$ & 6.207e+04 $+$ & 5.183e+04 \\
			\midrule
			\multirow{2}{*}{$f_4$} & 1.891e+03 $+$ & 4.183e+03 $+$ & 2.224e+03 $+$ & 8.573e+02 $+$ & 5.789e+02 $\approx$ & 1.638e+04 $+$ & 7.730e+03 $+$ & 8.936e+02 $+$ & 1.110e+03 $+$ & 8.852e+02 $+$ & 1.009e+04 $+$ & 4.823e+03 $+$ \\
                ~ & 1.256e+04 $+$ & 1.102e+04 $+$ & 1.064e+03 $+$ & \textbf{5.201e+02} $-$ & 1.694e+03 $+$ & 8.991e+03 $+$ & 1.220e+04 $+$ & 4.869e+03 $+$ & 5.642e+02 $\approx$ & 5.230e+02 $-$ & 5.482e+02 $\approx$ & 5.582e+02 \\
                
			\multirow{2}{*}{$f_5$} & 5.210e+02 $\approx$ & 5.209e+02 $\approx$ & 5.210e+02 $\approx$ & 5.210e+02 $\approx$ & 5.210e+02 $\approx$ & 5.209e+02 $\approx$ & 5.210e+02 $\approx$ & 5.210e+02 $\approx$ & \textbf{5.203e+02} $-$ & 5.207e+02 $-$ & 5.210e+02 $\approx$ & 5.209e+02 $\approx$ \\
                ~ & 5.213e+02 $+$ & 5.210e+02 $\approx$ & 5.208e+02 $\approx$ & 5.211e+02 $\approx$ & 5.205e+02 $-$ & 5.210e+02 $\approx$ & 5.211e+02 $\approx$ & 5.210e+02 $\approx$ & 5.210e+02 $\approx$ & 5.209e+02 $\approx$ & 5.210e+02 $\approx$ & 5.210e+02 \\
                
			\multirow{2}{*}{$f_6$} & 6.313e+02 $+$ & 6.370e+02 $+$ & 6.397e+02 $+$ & 6.379e+02 $+$ & \textbf{6.117e+02} $-$ & 6.387e+02 $+$ & 6.375e+02 $+$ & 6.253e+02 $\approx$ & 6.399e+02 $+$ & 6.378e+02 $+$ & 6.404e+02 $+$ & 6.314e+02 $+$ \\
                ~ & 6.445e+02 $+$ & 6.344e+02 $+$ & 6.260e+02 $\approx$ & 6.387e+02 $+$ & 6.210e+02 $+$ & 6.398e+02 $+$ & 6.337e+02 $+$ & 6.371e+02 $+$ & 6.283e+02 $\approx$ & 6.134e+02 $-$ & 6.274e+02 $\approx$ & 6.270e+02 \\
			\multirow{2}{*}{$f_7$} & 8.336e+02 $+$ & 1.029e+03 $+$ & 7.687e+02 $+$ & 7.284e+02 $+$ & 7.051e+02 $+$ & 1.381e+03 $+$ & 1.137e+03 $+$ & 7.357e+02 $+$ & 7.311e+02 $+$ & 7.153e+02 $+$ & 1.254e+03 $+$ & 1.011e+03 $+$ \\
                ~ & 1.418e+03 $+$ & 1.045e+03 $+$ & 8.171e+02 $+$ & 7.011e+02 $+$ & 8.165e+02 $+$ & 1.256e+03 $+$ & 1.033e+03 $+$ & 1.027e+03 $+$ & 7.012e+02 $+$ & \textbf{7.001e+02} $-$ & 7.011e+02 $+$ & 7.010e+02 \\
			\multirow{2}{*}{$f_8$} & 9.335e+02 $+$ & 1.109e+03 $+$ & 1.099e+03 $+$ & 1.059e+03 $+$ & 8.718e+02 $-$ & 1.212e+03 $+$ & 1.130e+03 $+$ & 9.810e+02 $+$ & 1.020e+03 $+$ & 9.607e+02 $+$ & 1.138e+03 $+$ & 1.012e+03 $+$ \\
                ~ & 1.256e+03 $+$ & 1.069e+03 $+$ & 1.031e+03 $+$ & 9.591e+02 $+$ & \textbf{8.675e+02} $-$ & 1.106e+03 $+$ & 1.086e+03 $+$ & 1.083e+03 $+$ & 9.281e+02 $+$ & 8.976e+02 $\approx$ & 9.202e+02 $\approx$ & 9.184e+02 \\
			\multirow{2}{*}{$f_9$} & 1.162e+03 $+$ & 1.285e+03 $+$ & 1.212e+03 $+$ & 1.170e+03 $+$ & \textbf{1.004e+03} $-$ & 1.428e+03 $+$ & 1.331e+03 $+$ & 1.109e+03 $+$ & 1.158e+03 $+$ & 1.083e+03 $+$ & 1.271e+03 $+$ & 1.145e+03 $+$ \\
                ~ & 1.303e+03 $+$ & 1.182e+03 $+$ & 1.137e+03 $+$ & 1.082e+03 $+$ & 1.026e+03 $\approx$ & 1.233e+03 $+$ & 1.215e+03 $+$ & 1.214e+03 $+$ & 1.061e+03 $+$ & 1.008e+03 $-$ & 1.044e+03 $\approx$ & 1.042e+03 \\
			\multirow{2}{*}{$f_{10}$} & 3.117e+03 $-$ & 6.513e+03 $+$ & 8.434e+03 $+$ & 7.939e+03 $+$ & 4.231e+03 $\approx$ & 7.607e+03 $+$ & 6.625e+03 $+$ & 6.053e+03 $+$ & 5.789e+03 $+$ & 5.518e+03 $+$ & 8.140e+03 $+$ & 6.953e+03 $+$ \\
                ~ & 9.731e+03 $+$ & 7.026e+03 $+$ & 6.128e+03 $+$ & 5.638e+03 $+$ & \textbf{2.468e+03} $-$ & 8.455e+03 $+$ & 7.637e+03 $+$ & 7.611e+03 $+$ & 4.494e+03 $+$ & 3.825e+03 $-$ & 4.412e+03 $+$ & 4.210e+03 \\
			\multirow{2}{*}{$f_{11}$} & 7.829e+03 $+$ & 8.636e+03 $+$ & 8.781e+03 $+$ & 8.567e+03 $+$ & 6.348e+03 $+$ & 8.020e+03 $+$ & 7.758e+03 $+$ & 7.458e+03 $+$ & 6.494e+03 $+$ & 6.289e+03 $+$ & 8.415e+03 $+$ & 7.659e+03 $+$ \\
                ~ & 9.910e+03 $+$ & 8.395e+03 $+$ & 6.794e+03 $+$ & 5.803e+03 $+$ & 4.913e+03 $\approx$ & 8.740e+03 $+$ & 8.378e+03 $+$ & 7.912e+03 $+$ & 5.341e+03 $+$ & \textbf{4.097e+03} $-$ & 5.106e+03 $\approx$ & 4.958e+03 \\
			\multirow{2}{*}{$f_{12}$} & 1.201e+03 $\approx$ & 1.203e+03 $+$ & 1.203e+03 $+$ & 1.203e+03 $+$ & 1.203e+03 $+$ & 1.202e+03 $+$ & 1.203e+03 $+$ & 1.203e+03 $+$ & 1.202e+03 $+$ & 1.202e+03 $+$ & 1.203e+03 $+$ & 1.202e+03 $+$ \\
                ~ & 1.204e+03 $+$ & 1.203e+03 $+$ & 1.201e+03 $\approx$ & 1.201e+03 $\approx$ & \textbf{1.201e+03} $-$ & 1.203e+03 $+$ & 1.203e+03 $+$ & 1.202e+03 $+$ & 1.202e+03 $+$ & 1.201e+03 $\approx$ & 1.202e+03 $+$ & 1.201e+03 \\
			\multirow{2}{*}{$f_{13}$} & 1.303e+03 $+$ & 1.305e+03 $+$ & 1.301e+03 $+$ & 1.301e+03 $+$ & \textbf{1.301e+03} $-$ & 1.308e+03 $+$ & 1.306e+03 $+$ & 1.301e+03 $\approx$ & 1.301e+03 $\approx$ & 1.301e+03 $\approx$ & 1.308e+03 $+$ & 1.305e+03 $+$ \\
                ~ & 1.307e+03 $+$ & 1.307e+03 $+$ & 1.302e+03 $+$ & 1.301e+03 $\approx$ & 1.302e+03 $+$ & 1.306e+03 $+$ & 1.307e+03 $+$ & 1.305e+03 $+$ & 1.301e+03 $\approx$ & 1.301e+03 $\approx$ & 1.301e+03 $\approx$ & 1.301e+03 \\
			\multirow{2}{*}{$f_{14}$} & 1.440e+03 $+$ & 1.511e+03 $+$ & 1.415e+03 $+$ & 1.408e+03 $+$ & 1.400e+03 $\approx$ & 1.632e+03 $+$ & 1.542e+03 $+$ & 1.407e+03 $+$ & 1.419e+03 $+$ & 1.404e+03 $+$ & 1.572e+03 $+$ & 1.493e+03 $+$ \\
                ~ & 1.617e+03 $+$ & 1.464e+03 $+$ & 1.431e+03 $+$ & 1.400e+03 $\approx$ & 1.433e+03 $+$ & 1.565e+03 $+$ & 1.452e+03 $+$ & 1.506e+03 $+$ & \textbf{1.400e+03} $\approx$ & 1.400e+03 $\approx$ & 1.400e+03 $\approx$ & 1.400e+03 \\
			\multirow{2}{*}{$f_{15}$} & 1.778e+04 $+$ & 1.820e+06 $+$ & 3.933e+03 $+$ & 3.668e+03 $+$ & 1.520e+03 $-$ & 1.446e+07 $+$ & 3.141e+06 $+$ & 1.607e+03 $+$ & 3.981e+03 $+$ & 1.599e+03 $+$ & 7.827e+04 $+$ & 4.659e+04 $+$ \\
                ~ & 4.524e+05 $+$ & 1.146e+04 $+$ & 6.950e+03 $+$ & 1.540e+03 $+$ & 1.054e+04 $+$ & 1.383e+05 $+$ & 5.256e+03 $+$ & 8.163e+04 $+$ & 1.527e+03 $\approx$ & \textbf{1.511e+03} $-$ & 1.528e+03 $\approx$ & 1.527e+03 \\
			\multirow{2}{*}{$f_{16}$} & 1.612e+03 $\approx$ & 1.614e+03 $+$ & 1.613e+03 $+$ & 1.613e+03 $+$ & 1.612e+03 $+$ & 1.614e+03 $+$ & 1.614e+03 $+$ & 1.613e+03 $+$ & 1.613e+03 $+$ & 1.613e+03 $+$ & 1.613e+03 $+$ & 1.613e+03 $+$ \\
                ~ & 1.614e+03 $+$ & 1.613e+03 $+$ & 1.613e+03 $\approx$ & 1.614e+03 $+$ & 1.613e+03 $+$ & 1.613e+03 $+$ & 1.613e+03 $+$ & 1.613e+03 $+$ & 1.613e+03 $\approx$ & \textbf{1.612e+03} $-$ & 1.613e+03 $\approx$ & 1.612e+03 \\
			\bottomrule
		\end{tabular}
	}
\end{sidewaystable}
\begin{sidewaystable}[htbp]
	\scriptsize
	\centering
	\renewcommand\arraystretch{1.5}
	\caption{Experimental and statistical results between ZSO\_Gauss and other competitor algorithms on 30-D CEC2014 benchmark functions (Continued).}
	\label{tbl:4.2.1.2}
	\resizebox{\columnwidth}{!}{
		\begin{tabular}{ccccccccccccccccccccccccc}
			\toprule
			\multirow{2}{*}{Func.} & GA & DE & PSO & CMA-ES & GWO & MFO & Jaya & SCA & WOA & HHO & AO & ArchOA \\
                ~ & AOA & SOA & GJO & FOX & EVO & SHO & CDO & COA & ZSO\_C & ZSO\_L & ZSO\_U & ZSO\_G \\
			\midrule
			\multirow{2}{*}{$f_{17}$} & 1.348e+07 $+$ & 1.418e+07 $+$ & 3.294e+07 $+$ & 7.038e+05 $\approx$ & 8.355e+05 $\approx$ & 5.807e+07 $+$ & 1.608e+07 $+$ & 4.540e+06 $+$ & 1.302e+07 $+$ & 2.181e+07 $+$ & 8.122e+07 $+$ & 1.958e+07 $+$ \\
                ~ & 2.565e+08 $+$ & 9.187e+07 $+$ & 1.027e+07 $+$ & \textbf{5.523e+05} $-$ & 1.616e+07 $+$ & 9.747e+07 $+$ & 3.161e+07 $+$ & 1.186e+07 $+$ & 1.730e+06 $+$ & 1.281e+06 $\approx$ & 1.416e+06 $\approx$ & 1.252e+06 \\
			\multirow{2}{*}{$f_{18}$} & 4.270e+08 $+$ & 4.359e+08 $+$ & 6.307e+08 $+$ & 1.765e+07 $+$ & 2.698e+05 $+$ & 5.397e+08 $+$ & 1.576e+08 $+$ & 1.258e+07 $+$ & 1.149e+05 $+$ & 1.953e+05 $+$ & 1.801e+09 $+$ & 2.423e+08 $+$ \\
                ~ & 8.633e+09 $+$ & 3.687e+09 $+$ & 2.763e+07 $+$ & 2.816e+05 $+$ & 9.478e+06 $+$ & 2.514e+09 $+$ & 2.619e+09 $+$ & 3.146e+08 $+$ & 4.879e+05 $+$ & 1.073e+05 $+$ & 1.465e+05 $+$ & \textbf{6.934e+04} \\
			\multirow{2}{*}{$f_{19}$} & 1.116e+07 $+$ & 9.463e+06 $+$ & 2.222e+07 $+$ & 4.433e+04 $+$ & 1.309e+04 $+$ & 1.310e+08 $+$ & 1.088e+07 $+$ & 2.537e+04 $+$ & 1.523e+04 $+$ & 1.102e+05 $+$ & 1.481e+08 $+$ & 7.584e+06 $+$ \\
                ~ & 8.143e+08 $+$ & 1.178e+08 $+$ & 4.887e+05 $+$ & 3.180e+03 $+$ & 6.401e+04 $+$ & 1.208e+08 $+$ & 6.874e+08 $+$ & 3.068e+06 $+$ & 3.098e+03 $+$ & \textbf{2.128e+03} $\approx$ & 3.333e+03 $+$ & 2.693e+03 \\
			\multirow{2}{*}{$f_{20}$} & 1.806e+10 $+$ & 1.990e+10 $+$ & 7.815e+11 $+$ & 1.839e+06 $+$ & 1.419e+05 $+$ & 8.473e+13 $+$ & 1.902e+12 $+$ & 3.579e+07 $+$ & 9.036e+08 $+$ & 1.374e+08 $+$ & 2.212e+13 $+$ & 5.872e+11 $+$ \\
                ~ & 6.919e+14 $+$ & 6.808e+10 $+$ & 6.005e+09 $+$ & 1.163e+05 $\approx$ & 3.132e+10 $+$ & 3.859e+13 $+$ & 8.233e+10 $+$ & 7.272e+10 $+$ & 8.898e+04 $\approx$ & 1.269e+05 $\approx$ & \textbf{8.857e+04} $\approx$ & 9.890e+04 \\
			\multirow{2}{*}{$f_{21}$} & 2.520e+07 $+$ & 2.636e+07 $+$ & 7.972e+07 $+$ & 6.877e+05 $\approx$ & 7.994e+05 $\approx$ & 5.522e+08 $+$ & 1.265e+08 $+$ & 5.693e+06 $+$ & 2.752e+06 $+$ & 1.074e+07 $+$ & 4.939e+08 $+$ & 6.107e+07 $+$ \\
                ~ & 2.358e+09 $+$ & 4.386e+08 $+$ & 4.353e+07 $+$ & \textbf{4.205e+05} $\approx$ & 1.417e+07 $+$ & 5.106e+08 $+$ & 1.571e+09 $+$ & 9.842e+07 $+$ & 7.447e+05 $\approx$ & 5.989e+05 $\approx$ & 6.055e+05 $\approx$ & 5.543e+05 \\
			\multirow{2}{*}{$f_{22}$} & 6.540e+07 $+$ & 7.416e+07 $+$ & 4.973e+09 $+$ & 4.472e+03 $+$ & 5.329e+03 $+$ & 2.914e+12 $+$ & 3.490e+09 $+$ & 5.555e+03 $+$ & 7.461e+03 $+$ & 6.939e+03 $+$ & 1.428e+11 $+$ & 2.547e+05 $+$ \\
                ~ & 1.242e+13 $+$ & 2.223e+11 $+$ & 2.674e+06 $+$ & \textbf{3.218e+03} $-$ & 6.987e+08 $+$ & 5.457e+10 $+$ & 1.852e+14 $+$ & 1.116e+06 $+$ & 4.231e+03 $\approx$ & 5.015e+03 $+$ & 4.239e+03 $\approx$ & 4.204e+03 \\
			\midrule
			\multirow{2}{*}{$f_{23}$} & 2.585e+03 $+$ & 2.727e+03 $+$ & 2.552e+03 $+$ & 2.554e+03 $+$ & 2.528e+03 $+$ & 3.354e+03 $+$ & 3.007e+03 $+$ & 2.543e+03 $+$ & 2.548e+03 $+$ & \textbf{2.500e+03} $-$ & \textbf{2.500e+03} $-$ & 2.588e+03 $+$ \\
                ~ & 2.512e+03 $-$ & 2.507e+03 $-$ & 2.543e+03 $\approx$ & \textbf{2.500e+03} $-$ & 2.560e+03 $+$ & 2.533e+03 $-$ & \textbf{2.500e+03} $-$ & \textbf{2.500e+03} $-$ & 2.540e+03 $\approx$ & 2.540e+03 $\approx$ & 2.541e+03 $\approx$ & 2.541e+03 \\
			\multirow{2}{*}{$f_{24}$} & 2.701e+03 $+$ & 2.891e+03 $+$ & 2.654e+03 $+$ & 2.678e+03 $+$ & 2.601e+03 $-$ & 2.980e+03 $+$ & 2.907e+03 $+$ & 2.601e+03 $-$ & \textbf{2.600e+03} $-$ & \textbf{2.600e+03} $-$ & 2.601e+03 $-$ & 2.601e+03 $-$ \\
                ~ & 2.602e+03 $-$ & \textbf{2.600e+03} $-$ & \textbf{2.600e+03} $-$ & \textbf{2.600e+03} $-$ & 2.647e+03 $+$ & \textbf{2.600e+03} $-$ & \textbf{2.600e+03} $-$ & \textbf{2.600e+03} $-$ & 2.627e+03 $\approx$ & 2.602e+03 $-$ & 2.625e+03 $\approx$ & 2.623e+03 \\
			\multirow{2}{*}{$f_{25}$} & 2.750e+03 $+$ & 2.754e+03 $+$ & 2.758e+03 $+$ & 2.713e+03 $\approx$ & 2.715e+03 $\approx$ & 2.834e+03 $+$ & 2.797e+03 $+$ & 2.719e+03 $\approx$ & 2.723e+03 $+$ & \textbf{2.700e+03} $-$ & \textbf{2.700e+03} $-$ & \textbf{2.700e+03} $-$ \\
                ~ & 2.701e+03 $-$ & \textbf{2.700e+03} $-$ & 2.702e+03 $-$ & \textbf{2.700e+03} $-$ & 2.719e+03 $+$ & \textbf{2.700e+03} $-$ & \textbf{2.700e+03} $-$ & \textbf{2.700e+03} $-$ & 2.714e+03 $\approx$ & 2.705e+03 $\approx$ & 2.715e+03 $\approx$ & 2.714e+03 \\
			\multirow{2}{*}{$f_{26}$} & 2.703e+03 $\approx$ & 2.705e+03 $\approx$ & 2.703e+03 $\approx$ & 2.701e+03 $-$ & 2.737e+03 $\approx$ & 2.709e+03 $\approx$ & 2.705e+03 $\approx$ & \textbf{2.700e+03} $-$ & 2.702e+03 $\approx$ & 2.748e+03 $\approx$ & 2.715e+03 $\approx$ & 2.709e+03 $\approx$ \\
                ~ & 2.789e+03 $\approx$ & 2.737e+03 $\approx$ & 2.741e+03 $\approx$ & 2.783e+03 $\approx$ & 2.732e+03 $\approx$ & 2.761e+03 $\approx$ & 2.728e+03 $\approx$ & 2.707e+03 $\approx$ & 2.764e+03 $\approx$ & 2.740e+03 $\approx$ & 2.744e+03 $\approx$ & 2.747e+03 \\
			\multirow{2}{*}{$f_{27}$} & 3.819e+03 $\approx$ & 3.968e+03 $+$ & 3.850e+03 $+$ & 3.935e+03 $+$ & 3.309e+03 $-$ & 4.005e+03 $+$ & 3.983e+03 $+$ & 3.549e+03 $-$ & 4.030e+03 $+$ & \textbf{2.900e+03} $-$ & 4.065e+03 $+$ & 3.948e+03 $+$ \\
                ~ & 2.903e+03 $-$ & 3.952e+03 $+$ & 3.685e+03 $\approx$ & \textbf{2.900e+03} $-$ & 3.656e+03 $\approx$ & 4.227e+03 $+$ & \textbf{2.900e+03} $-$ & \textbf{2.900e+03} $-$ & 3.752e+03 $\approx$ & 3.451e+03 $-$ & 3.717e+03 $\approx$ & 3.741e+03 \\
			\multirow{2}{*}{$f_{28}$} & 4.452e+03 $-$ & 3.967e+03 $-$ & 6.733e+03 $+$ & 4.009e+03 $-$ & 3.821e+03 $-$ & 5.169e+03 $-$ & 4.210e+03 $-$ & 5.078e+03 $-$ & 5.834e+03 $\approx$ & \textbf{3.000e+03} $-$ & 5.349e+03 $-$ & 5.076e+03 $-$ \\
                ~ & 3.010e+03 $-$ & 6.600e+03 $\approx$ & 4.467e+03 $-$ & \textbf{3.000e+03} $-$ & 4.454e+03 $-$ & 8.603e+03 $+$ & \textbf{3.000e+03} $-$ & \textbf{3.000e+03} $-$ & 6.904e+03 $+$ & 4.698e+03 $-$ & 6.693e+03 $+$ & 6.279e+03 \\
			\multirow{2}{*}{$f_{29}$} & 2.331e+08 $+$ & 8.159e+07 $+$ & 3.370e+08 $+$ & 6.718e+06 $+$ & 7.682e+06 $+$ & 5.132e+08 $+$ & 2.460e+08 $+$ & 3.644e+07 $+$ & 5.417e+07 $+$ & \textbf{3.108e+03} $-$ & 6.600e+06 $+$ & 5.184e+07 $+$ \\
                ~ & 1.256e+07 $+$ & \textbf{3.108e+03} $-$ & 1.065e+08 $+$ & \textbf{3.108e+03} $-$ & 2.332e+08 $+$ & 8.891e+06 $+$ & \textbf{3.108e+03} $-$ & \textbf{3.108e+03} $-$ & 5.735e+06 $+$ & 4.068e+06 $+$ & 5.064e+06 $+$ & 2.168e+06 \\
			\multirow{2}{*}{$f_{30}$} & 2.315e+09 $+$ & 5.148e+08 $+$ & 8.188e+09 $+$ & 4.248e+07 $+$ & 2.760e+07 $+$ & 4.517e+11 $+$ & 7.932e+09 $+$ & 5.721e+08 $+$ & 4.156e+08 $+$ & 6.393e+08 $+$ & 5.190e+09 $+$ & 3.207e+09 $+$ \\
                ~ & 1.979e+11 $+$ & 7.132e+10 $+$ & 4.562e+08 $+$ & 1.675e+05 $+$ & 3.555e+09 $+$ & 5.492e+10 $+$ & \textbf{3.205e+03} $-$ & 6.044e+08 $+$ & 5.016e+07 $+$ & 1.056e+07 $\approx$ & 2.929e+07 $+$ & 1.838e+07 \\
                \midrule
                \multirow{2}{*}{$+$/$\approx$/$-$:} & 22/5/3 & 27/2/1 & 28/2/0 & 22/4/4 & 13/8/9 & 27/2/1 & 27/1/2 & 21/5/4 & 24/4/2 & 20/3/7 & 24/2/4 & 25/2/3 \\
                ~ & 24/1/5 & 23/3/4 & 20/7/3 & 13/7/10 & 20/4/6 & 25/2/3 & 21/2/7 & 22/2/6 & 14/16/0 & 4/14/12 & 10/20/0 & - \\
                \midrule
                \multirow{2}{*}{Ave ranks:} & 12.4 & 15.8 & 17.1 & 10.9 & 6.7 & 20.5 & 18.1 & 10.0 & 10.9 & 8.0 & 18.0 & 13.7 \\
                ~ & 20.6 & 16.1 & 11.0 & 6.9 & 10.6 & 18.3 & 14.5 & 11.9 & 8.7 & \textbf{4.9} & 7.6 & 6.2 \\
			\bottomrule
		\end{tabular}
	}
\end{sidewaystable}
\begin{sidewaystable}[htbp]
	\scriptsize
	\centering
	\renewcommand\arraystretch{1.5}
	\caption{Experimental and statistical results between ZSO\_Gauss and other competitor algorithms on 50-D CEC2014 benchmark functions.}
	\label{tbl:4.2.1.3}
	\resizebox{\columnwidth}{!}{
		\begin{tabular}{ccccccccccccccccccccccccc}
			\toprule
			\multirow{2}{*}{Func.} & GA & DE & PSO & CMA-ES & GWO & MFO & Jaya & SCA & WOA & HHO & AO & ArchOA \\
                ~ & AOA & SOA & GJO & FOX & EVO & SHO & CDO & COA & ZSO\_C & ZSO\_L & ZSO\_U & ZSO\_G \\
			\midrule
			\multirow{2}{*}{$f_1$} & 1.098e+09 $+$ & 1.260e+09 $+$ & 9.492e+08 $+$ & 7.728e+07 $+$ & 6.181e+07 $+$ & 2.652e+09 $+$ & 1.423e+09 $+$ & 2.425e+08 $+$ & 1.391e+08 $+$ & 1.631e+08 $+$ & 3.541e+09 $+$ & 1.111e+09 $+$ \\
                ~ & 5.594e+09 $+$ & 1.698e+09 $+$ & 4.915e+08 $+$ & \textbf{1.532e+07} $-$ & 4.415e+08 $+$ & 3.015e+09 $+$ & 1.621e+09 $+$ & 1.419e+09 $+$ & 2.279e+07 $-$ & 2.732e+07 $\approx$ & 2.725e+07 $\approx$ & 2.943e+07 \\
			\multirow{2}{*}{$f_2$} & 5.705e+10 $+$ & 1.025e+11 $+$ & 2.763e+10 $+$ & 1.220e+10 $+$ & 3.379e+09 $+$ & 1.954e+11 $+$ & 1.299e+11 $+$ & 1.307e+10 $+$ & 2.732e+10 $+$ & 4.659e+09 $+$ & 1.372e+11 $+$ & 1.019e+11 $+$ \\
                ~ & 1.578e+11 $+$ & 1.469e+11 $+$ & 4.733e+10 $+$ & 3.460e+07 $+$ & 4.357e+10 $+$ & 1.441e+11 $+$ & 1.237e+11 $+$ & 1.052e+11 $+$ & 4.344e+07 $+$ & \textbf{6.923e+04} $-$ & 1.299e+07 $\approx$ & 1.299e+07 \\
			\multirow{2}{*}{$f_3$} & 2.005e+05 $+$ & 3.220e+05 $+$ & 1.995e+05 $+$ & \textbf{2.237e+04} $-$ & 3.450e+04 $-$ & 4.424e+05 $+$ & 4.090e+05 $+$ & 1.252e+05 $+$ & 9.394e+04 $+$ & 8.675e+04 $+$ & 3.346e+05 $+$ & 1.416e+05 $+$ \\
                ~ & 5.854e+05 $+$ & 1.387e+05 $+$ & 1.311e+05 $+$ & 1.354e+05 $+$ & 1.686e+05 $+$ & 1.573e+05 $+$ & 1.562e+05 $+$ & 1.534e+05 $+$ & 1.159e+05 $+$ & 9.776e+04 $+$ & 9.890e+04 $+$ & 7.875e+04 \\
			\midrule
			\multirow{2}{*}{$f_4$} & 9.309e+03 $+$ & 1.956e+04 $+$ & 7.954e+03 $+$ & 1.669e+03 $+$ & 8.150e+02 $+$ & 6.257e+04 $+$ & 2.414e+04 $+$ & 2.393e+03 $+$ & 3.173e+03 $+$ & 1.393e+03 $+$ & 3.607e+04 $+$ & 2.159e+04 $+$ \\
                ~ & 4.573e+04 $+$ & 1.948e+04 $+$ & 6.322e+03 $+$ & 5.749e+02 $-$ & 7.335e+03 $+$ & 3.523e+04 $+$ & 1.097e+04 $+$ & 2.517e+04 $+$ & 6.383e+02 $\approx$ & \textbf{5.727e+02} $-$ & 6.723e+02 $\approx$ & 6.556e+02 \\
			\multirow{2}{*}{$f_5$} & 5.212e+02 $\approx$ & 5.212e+02 $\approx$ & 5.212e+02 $\approx$ & 5.212e+02 $\approx$ & 5.212e+02 $\approx$ & 5.210e+02 $\approx$ & 5.212e+02 $\approx$ & 5.212e+02 $\approx$ & \textbf{5.203e+02} $-$ & 5.209e+02 $-$ & 5.212e+02 $\approx$ & 5.211e+02 $\approx$ \\
                ~ & 5.214e+02 $+$ & 5.212e+02 $\approx$ & 5.210e+02 $-$ & 5.212e+02 $\approx$ & 5.206e+02 $-$ & 5.212e+02 $\approx$ & 5.212e+02 $\approx$ & 5.211e+02 $\approx$ & 5.212e+02 $\approx$ & 5.211e+02 $\approx$ & 5.212e+02 $\approx$ & 5.212e+02 \\
			\multirow{2}{*}{$f_6$} & 6.650e+02 $+$ & 6.638e+02 $+$ & 6.729e+02 $+$ & 6.599e+02 $+$ & 6.271e+02 $-$ & 6.683e+02 $+$ & 6.636e+02 $+$ & 6.509e+02 $\approx$ & 6.721e+02 $+$ & 6.695e+02 $+$ & 6.745e+02 $+$ & 6.622e+02 $+$ \\
                ~ & 6.795e+02 $+$ & 6.655e+02 $+$ & 6.578e+02 $+$ & 6.670e+02 $+$ & 6.449e+02 $-$ & 6.731e+02 $+$ & 6.646e+02 $+$ & 6.679e+02 $+$ & 6.551e+02 $+$ & \textbf{6.257e+02} $-$ & 6.550e+02 $+$ & 6.526e+02 \\
			\multirow{2}{*}{$f_7$} & 1.238e+03 $+$ & 1.772e+03 $+$ & 9.529e+02 $+$ & 8.238e+02 $+$ & 7.247e+02 $+$ & 2.684e+03 $+$ & 2.008e+03 $+$ & 8.096e+02 $+$ & 9.274e+02 $+$ & 7.463e+02 $+$ & 2.104e+03 $+$ & 1.612e+03 $+$ \\
                ~ & 2.183e+03 $+$ & 1.724e+03 $+$ & 1.105e+03 $+$ & 7.014e+02 $+$ & 1.127e+03 $+$ & 2.022e+03 $+$ & 1.474e+03 $+$ & 1.727e+03 $+$ & 7.014e+02 $+$ & \textbf{7.001e+02} $-$ & 7.012e+02 $\approx$ & 7.011e+02 \\
			\multirow{2}{*}{$f_8$} & 1.179e+03 $+$ & 1.416e+03 $+$ & 1.373e+03 $+$ & 1.289e+03 $+$ & \textbf{9.724e+02} $-$ & 1.607e+03 $+$ & 1.437e+03 $+$ & 1.184e+03 $+$ & 1.242e+03 $+$ & 1.095e+03 $+$ & 1.446e+03 $+$ & 1.266e+03 $+$ \\
                ~ & 1.604e+03 $+$ & 1.350e+03 $+$ & 1.315e+03 $+$ & 1.098e+03 $+$ & 9.904e+02 $-$ & 1.414e+03 $+$ & 1.388e+03 $+$ & 1.375e+03 $+$ & 1.076e+03 $+$ & 1.014e+03 $-$ & 1.065e+03 $\approx$ & 1.056e+03 \\
			\multirow{2}{*}{$f_9$} & 1.486e+03 $+$ & 1.720e+03 $+$ & 1.446e+03 $+$ & 1.412e+03 $+$ & \textbf{1.105e+03} $-$ & 1.992e+03 $+$ & 1.765e+03 $+$ & 1.324e+03 $+$ & 1.412e+03 $+$ & 1.326e+03 $+$ & 1.668e+03 $+$ & 1.429e+03 $+$ \\
                ~ & 1.707e+03 $+$ & 1.523e+03 $+$ & 1.411e+03 $+$ & 1.308e+03 $+$ & 1.200e+03 $\approx$ & 1.597e+03 $+$ & 1.494e+03 $+$ & 1.549e+03 $+$ & 1.244e+03 $+$ & 1.120e+03 $-$ & 1.208e+03 $\approx$ & 1.220e+03 \\
			\multirow{2}{*}{$f_{10}$} & 7.887e+03 $+$ & 1.264e+04 $+$ & 1.509e+04 $+$ & 1.475e+04 $+$ & 6.772e+03 $-$ & 1.401e+04 $+$ & 1.288e+04 $+$ & 1.161e+04 $+$ & 1.034e+04 $+$ & 9.227e+03 $+$ & 1.406e+04 $+$ & 1.254e+04 $+$ \\
                ~ & 1.681e+04 $+$ & 1.307e+04 $+$ & 1.232e+04 $+$ & 9.397e+03 $+$ & \textbf{4.509e+03} $-$ & 1.504e+04 $+$ & 1.369e+04 $+$ & 1.388e+04 $+$ & 7.842e+03 $+$ & 6.488e+03 $-$ & 7.591e+03 $\approx$ & 7.461e+03 \\
			\multirow{2}{*}{$f_{11}$} & 1.484e+04 $+$ & 1.511e+04 $+$ & 1.525e+04 $+$ & 1.511e+04 $+$ & 1.105e+04 $+$ & 1.402e+04 $+$ & 1.455e+04 $+$ & 1.315e+04 $+$ & 1.171e+04 $+$ & 1.056e+04 $+$ & 1.538e+04 $+$ & 1.374e+04 $+$ \\
                ~ & 1.684e+04 $+$ & 1.473e+04 $+$ & 1.331e+04 $+$ & 9.524e+03 $+$ & 9.219e+03 $+$ & 1.545e+04 $+$ & 1.486e+04 $+$ & 1.421e+04 $+$ & 8.429e+03 $\approx$ & \textbf{6.815e+03} $-$ & 8.143e+03 $\approx$ & 8.329e+03 \\
			\multirow{2}{*}{$f_{12}$} & 1.204e+03 $+$ & 1.204e+03 $+$ & 1.204e+03 $+$ & 1.204e+03 $+$ & 1.204e+03 $+$ & 1.202e+03 $\approx$ & 1.203e+03 $+$ & 1.203e+03 $+$ & 1.203e+03 $+$ & 1.203e+03 $+$ & 1.204e+03 $+$ & 1.203e+03 $+$ \\
                ~ & 1.206e+03 $+$ & 1.204e+03 $+$ & 1.202e+03 $+$ & 1.202e+03 $\approx$ & \textbf{1.201e+03} $-$ & 1.204e+03 $+$ & 1.204e+03 $+$ & 1.203e+03 $+$ & 1.202e+03 $\approx$ & 1.202e+03 $\approx$ & 1.202e+03 $\approx$ & 1.202e+03 \\
			\multirow{2}{*}{$f_{13}$} & 1.305e+03 $+$ & 1.306e+03 $+$ & 1.303e+03 $+$ & 1.301e+03 $+$ & 1.301e+03 $\approx$ & 1.309e+03 $+$ & 1.307e+03 $+$ & 1.301e+03 $+$ & 1.303e+03 $+$ & \textbf{1.301e+03} $-$ & 1.309e+03 $+$ & 1.307e+03 $+$ \\
                ~ & 1.309e+03 $+$ & 1.308e+03 $+$ & 1.304e+03 $+$ & 1.301e+03 $+$ & 1.304e+03 $+$ & 1.308e+03 $+$ & 1.308e+03 $+$ & 1.307e+03 $+$ & 1.301e+03 $\approx$ & 1.301e+03 $-$ & 1.301e+03 $\approx$ & 1.301e+03 \\
			\multirow{2}{*}{$f_{14}$} & 1.546e+03 $+$ & 1.708e+03 $+$ & 1.454e+03 $+$ & 1.430e+03 $+$ & 1.402e+03 $+$ & 1.936e+03 $+$ & 1.732e+03 $+$ & 1.429e+03 $+$ & 1.444e+03 $+$ & 1.407e+03 $+$ & 1.713e+03 $+$ & 1.620e+03 $+$ \\
                ~ & 1.754e+03 $+$ & 1.562e+03 $+$ & 1.500e+03 $+$ & 1.401e+03 $\approx$ & 1.491e+03 $+$ & 1.718e+03 $+$ & 1.523e+03 $+$ & 1.638e+03 $+$ & \textbf{1.400e+03} $\approx$ & 1.401e+03 $\approx$ & 1.401e+03 $\approx$ & 1.401e+03 \\
			\multirow{2}{*}{$f_{15}$} & 6.500e+05 $+$ & 1.088e+07 $+$ & 6.961e+04 $+$ & 2.304e+04 $+$ & 1.639e+03 $\approx$ & 1.186e+08 $+$ & 2.290e+07 $+$ & 6.644e+03 $+$ & 5.087e+04 $+$ & 3.213e+03 $+$ & 2.504e+06 $+$ & 8.125e+05 $+$ \\
                ~ & 5.696e+06 $+$ & 1.932e+06 $+$ & 1.092e+05 $+$ & 1.567e+03 $+$ & 1.469e+05 $+$ & 2.799e+06 $+$ & 1.337e+06 $+$ & 1.870e+06 $+$ & 1.553e+03 $\approx$ & \textbf{1.525e+03} $-$ & 1.557e+03 $\approx$ & 1.560e+03 \\
			\multirow{2}{*}{$f_{16}$} & 1.623e+03 $+$ & 1.623e+03 $+$ & 1.623e+03 $+$ & 1.623e+03 $+$ & 1.621e+03 $\approx$ & 1.624e+03 $+$ & 1.624e+03 $+$ & 1.622e+03 $+$ & 1.622e+03 $+$ & 1.622e+03 $+$ & 1.623e+03 $+$ & 1.622e+03 $+$ \\
                ~ & 1.624e+03 $+$ & 1.623e+03 $+$ & 1.622e+03 $+$ & 1.623e+03 $+$ & 1.623e+03 $+$ & 1.623e+03 $+$ & 1.623e+03 $+$ & 1.622e+03 $+$ & 1.622e+03 $\approx$ & \textbf{1.621e+03} $-$ & 1.622e+03 $\approx$ & 1.622e+03 \\
			\bottomrule
		\end{tabular}
	}
\end{sidewaystable}
\begin{sidewaystable}[htbp]
	\scriptsize
	\centering
	\renewcommand\arraystretch{1.5}
	\caption{Experimental and statistical results between ZSO\_Gauss and other competitor algorithms on 50-D CEC2014 benchmark functions (Continued).}
	\label{tbl:4.2.1.4}
	\resizebox{\columnwidth}{!}{
		\begin{tabular}{ccccccccccccccccccccccccc}
			\toprule
			\multirow{2}{*}{Func.} & GA & DE & PSO & CMA-ES & GWO & MFO & Jaya & SCA & WOA & HHO & AO & ArchOA \\
                ~ & AOA & SOA & GJO & FOX & EVO & SHO & CDO & COA & ZSO\_C & ZSO\_L & ZSO\_U & ZSO\_G \\
			\midrule
			\multirow{2}{*}{$f_{17}$} & 8.972e+07 $+$ & 8.506e+07 $+$ & 9.501e+07 $+$ & 6.540e+06 $+$ & 2.974e+06 $+$ & 1.718e+08 $+$ & 3.966e+07 $+$ & 1.235e+07 $+$ & 1.011e+07 $+$ & 2.384e+07 $+$ & 2.633e+08 $+$ & 5.587e+07 $+$ \\
                ~ & 6.968e+08 $+$ & 1.486e+08 $+$ & 3.130e+07 $+$ & \textbf{1.479e+06} $-$ & 3.011e+07 $+$ & 2.765e+08 $+$ & 4.305e+07 $+$ & 5.150e+07 $+$ & 2.738e+06 $\approx$ & 2.733e+06 $\approx$ & 2.489e+06 $\approx$ & 2.326e+06 \\
			\multirow{2}{*}{$f_{18}$} & 2.616e+09 $+$ & 2.175e+09 $+$ & 2.250e+09 $+$ & 1.238e+08 $+$ & 4.064e+05 $+$ & 3.001e+09 $+$ & 1.901e+09 $+$ & 7.758e+07 $+$ & 2.147e+06 $+$ & 6.454e+05 $+$ & 9.229e+09 $+$ & 2.299e+09 $+$ \\
                ~ & 1.232e+10 $+$ & 8.397e+09 $+$ & 4.826e+08 $+$ & 1.060e+06 $+$ & 1.293e+08 $+$ & 7.064e+09 $+$ & 1.233e+09 $+$ & 1.985e+09 $+$ & 1.982e+06 $+$ & 1.409e+05 $+$ & 3.794e+05 $+$ & \textbf{1.173e+05} \\
			\multirow{2}{*}{$f_{19}$} & 4.017e+08 $+$ & 3.123e+08 $+$ & 5.817e+08 $+$ & 1.267e+06 $+$ & 2.530e+04 $+$ & 1.685e+09 $+$ & 3.371e+08 $+$ & 1.434e+06 $+$ & 9.850e+04 $+$ & 1.977e+05 $+$ & 4.749e+09 $+$ & 5.801e+08 $+$ \\
                ~ & 7.336e+09 $+$ & 4.178e+09 $+$ & 1.873e+07 $+$ & 3.263e+03 $\approx$ & 1.056e+07 $+$ & 3.303e+09 $+$ & 5.861e+08 $+$ & 2.243e+08 $+$ & 3.671e+03 $\approx$ & 3.679e+03 $\approx$ & \textbf{2.940e+03} $\approx$ & 3.928e+03 \\
			\multirow{2}{*}{$f_{20}$} & 9.853e+12 $+$ & 8.214e+12 $+$ & 3.026e+13 $+$ & 9.329e+08 $+$ & 1.514e+07 $+$ & 9.031e+14 $+$ & 4.777e+12 $+$ & 8.246e+08 $+$ & 1.046e+08 $+$ & 5.761e+07 $+$ & 4.529e+14 $+$ & 2.397e+13 $+$ \\
                ~ & 1.457e+15 $+$ & 1.513e+12 $+$ & 7.176e+10 $+$ & 1.889e+05 $\approx$ & 4.705e+10 $+$ & 3.320e+14 $+$ & 1.090e+11 $+$ & 1.317e+13 $+$ & 1.605e+05 $\approx$ & 1.836e+05 $+$ & \textbf{1.484e+05} $\approx$ & 1.679e+05 \\
			\multirow{2}{*}{$f_{21}$} & 3.515e+08 $+$ & 3.167e+08 $+$ & 3.918e+08 $+$ & 6.854e+06 $+$ & 3.977e+06 $+$ & 1.442e+09 $+$ & 8.649e+08 $+$ & 2.254e+07 $+$ & 1.224e+07 $+$ & 1.480e+07 $+$ & 2.107e+09 $+$ & 3.702e+08 $+$ \\
                ~ & 3.189e+09 $+$ & 8.222e+08 $+$ & 1.388e+08 $+$ & \textbf{7.269e+05} $-$ & 9.609e+07 $+$ & 1.573e+09 $+$ & 7.524e+07 $+$ & 3.582e+08 $+$ & 8.631e+05 $\approx$ & 1.367e+06 $\approx$ & 8.258e+05 $\approx$ & 1.051e+06 \\
			\multirow{2}{*}{$f_{22}$} & 2.191e+11 $+$ & 2.835e+11 $+$ & 8.372e+11 $+$ & 3.008e+06 $+$ & 6.743e+03 $+$ & 5.480e+13 $+$ & 2.764e+11 $+$ & 8.788e+05 $+$ & 7.973e+03 $+$ & 7.411e+03 $+$ & 2.585e+13 $+$ & 1.039e+12 $+$ \\
                ~ & 2.296e+14 $+$ & 8.624e+12 $+$ & 2.423e+09 $+$ & \textbf{3.003e+03} $-$ & 1.478e+10 $+$ & 1.192e+13 $+$ & 1.088e+10 $+$ & 3.246e+10 $+$ & 5.643e+03 $\approx$ & 6.119e+03 $+$ & 5.156e+03 $\approx$ & 5.407e+03 \\
			\midrule
			\multirow{2}{*}{$f_{23}$} & 2.674e+03 $+$ & 3.065e+03 $+$ & 2.553e+03 $+$ & 2.560e+03 $+$ & 2.536e+03 $\approx$ & 4.279e+03 $+$ & 3.814e+03 $+$ & 2.538e+03 $\approx$ & 2.534e+03 $\approx$ & \textbf{2.500e+03} $-$ & \textbf{2.500e+03} $-$ & 2.543e+03 $\approx$ \\
                ~ & 2.506e+03 $-$ & \textbf{2.500e+03} $-$ & 2.540e+03 $\approx$ & \textbf{2.500e+03} $-$ & 2.565e+03 $+$ & \textbf{2.500e+03} $-$ & \textbf{2.500e+03} $-$ & \textbf{2.500e+03} $-$ & 2.536e+03 $\approx$ & 2.535e+03 $\approx$ & 2.536e+03 $\approx$ & 2.536e+03 \\
			\multirow{2}{*}{$f_{24}$} & 2.888e+03 $+$ & 3.248e+03 $+$ & 2.740e+03 $+$ & 2.758e+03 $+$ & 2.601e+03 $-$ & 3.384e+03 $+$ & 3.266e+03 $+$ & 2.601e+03 $-$ & 2.606e+03 $-$ & \textbf{2.600e+03} $-$ & 2.601e+03 $-$ & 2.601e+03 $-$ \\
                ~ & 2.604e+03 $-$ & \textbf{2.600e+03} $-$ & \textbf{2.600e+03} $-$ & \textbf{2.600e+03} $-$ & 2.719e+03 $+$ & \textbf{2.600e+03} $-$ & \textbf{2.600e+03} $-$ & \textbf{2.600e+03} $-$ & 2.701e+03 $\approx$ & 2.665e+03 $-$ & 2.697e+03 $\approx$ & 2.696e+03 \\
			\multirow{2}{*}{$f_{25}$} & 2.850e+03 $+$ & 2.816e+03 $+$ & 2.813e+03 $+$ & 2.733e+03 $\approx$ & 2.727e+03 $\approx$ & 2.962e+03 $+$ & 2.856e+03 $+$ & 2.720e+03 $\approx$ & 2.703e+03 $-$ & \textbf{2.700e+03} $-$ & \textbf{2.700e+03} $-$ & \textbf{2.700e+03} $-$ \\
                ~ & 2.701e+03 $-$ & \textbf{2.700e+03} $-$ & \textbf{2.700e+03} $-$ & \textbf{2.700e+03} $-$ & 2.742e+03 $+$ & \textbf{2.700e+03} $-$ & \textbf{2.700e+03} $-$ & \textbf{2.700e+03} $-$ & 2.729e+03 $\approx$ & 2.703e+03 $-$ & 2.727e+03 $\approx$ & 2.728e+03 \\
			\multirow{2}{*}{$f_{26}$} & 2.817e+03 $+$ & 2.707e+03 $-$ & 2.705e+03 $-$ & 2.725e+03 $-$ & 2.794e+03 $\approx$ & 2.711e+03 $-$ & 2.744e+03 $-$ & 2.737e+03 $-$ & \textbf{2.703e+03} $-$ & 2.770e+03 $-$ & 2.725e+03 $\approx$ & 2.731e+03 $-$ \\
                ~ & 2.800e+03 $+$ & 2.710e+03 $-$ & 2.790e+03 $\approx$ & 2.800e+03 $+$ & 2.787e+03 $\approx$ & 2.795e+03 $\approx$ & 2.706e+03 $-$ & 2.712e+03 $-$ & 2.794e+03 $\approx$ & 2.787e+03 $\approx$ & 2.801e+03 $\approx$ & 2.794e+03 \\
			\multirow{2}{*}{$f_{27}$} & 4.688e+03 $+$ & 4.848e+03 $+$ & 4.826e+03 $+$ & 4.697e+03 $+$ & 3.602e+03 $-$ & 4.811e+03 $+$ & 4.710e+03 $+$ & 4.294e+03 $-$ & 5.109e+03 $+$ & \textbf{2.900e+03} $-$ & 4.961e+03 $+$ & 4.739e+03 $+$ \\
                ~ & 3.479e+03 $-$ & 4.950e+03 $+$ & 4.496e+03 $\approx$ & \textbf{2.900e+03} $-$ & 4.341e+03 $-$ & 5.280e+03 $+$ & \textbf{2.900e+03} $-$ & \textbf{2.900e+03} $-$ & 4.641e+03 $\approx$ & 3.970e+03 $-$ & 4.687e+03 $\approx$ & 4.570e+03 \\
			\multirow{2}{*}{$f_{28}$} & 7.187e+03 $-$ & 4.744e+03 $-$ & 1.272e+04 $+$ & 5.044e+03 $-$ & 4.827e+03 $-$ & 7.252e+03 $-$ & 4.889e+03 $-$ & 9.603e+03 $-$ & 1.014e+04 $\approx$ & \textbf{3.000e+03} $-$ & 8.205e+03 $-$ & 8.662e+03 $-$ \\
                ~ & 3.043e+03 $-$ & 1.199e+04 $+$ & 8.240e+03 $-$ & \textbf{3.000e+03} $-$ & 6.639e+03 $-$ & 1.536e+04 $+$ & \textbf{3.000e+03} $-$ & \textbf{3.000e+03} $-$ & 1.117e+04 $+$ & 6.625e+03 $-$ & 1.105e+04 $\approx$ & 1.027e+04 \\
			\multirow{2}{*}{$f_{29}$} & 1.946e+09 $+$ & 4.243e+08 $+$ & 1.730e+09 $+$ & 5.952e+07 $+$ & 2.491e+07 $+$ & 2.750e+09 $+$ & 1.173e+09 $+$ & 1.506e+08 $+$ & 5.075e+07 $+$ & \textbf{3.114e+03} $-$ & 1.070e+05 $-$ & \textbf{3.114e+03} $-$ \\
                ~ & 1.646e+07 $\approx$ & \textbf{3.114e+03} $-$ & 3.264e+08 $+$ & \textbf{3.114e+03} $-$ & 6.865e+08 $+$ & \textbf{3.114e+03} $-$ & \textbf{3.114e+03} $-$ & \textbf{3.114e+03} $-$ & 7.254e+06 $\approx$ & 4.201e+05 $-$ & 6.827e+06 $\approx$ & 1.720e+07 \\
			\multirow{2}{*}{$f_{30}$} & 8.436e+10 $+$ & 1.305e+10 $+$ & 1.455e+11 $+$ & 2.454e+08 $+$ & 1.320e+08 $+$ & 1.873e+13 $+$ & 2.294e+11 $+$ & 1.537e+09 $+$ & 1.288e+09 $+$ & 2.931e+07 $+$ & 4.915e+09 $+$ & 1.503e+10 $+$ \\
                ~ & 2.668e+10 $+$ & 2.748e+10 $+$ & 1.486e+09 $+$ & \textbf{3.207e+03} $-$ & 5.053e+09 $+$ & 7.306e+10 $+$ & \textbf{3.207e+03} $-$ & \textbf{3.207e+03} $-$ & 1.450e+07 $\approx$ & 1.112e+07 $\approx$ & 1.194e+07 $\approx$ & 1.364e+07 \\
                \midrule
                \multirow{2}{*}{$+$/$\approx$/$-$:} & 28/1/1 & 27/1/2 & 28/1/1 & 25/2/3 & 15/7/8 & 26/2/2 & 27/1/2 & 22/4/4 & 24/2/4 & 21/0/9 & 23/2/5 & 23/2/5 \\
                ~ & 24/1/5 & 24/1/5 & 23/3/4 & 13/5/12 & 21/2/7 & 24/2/4 & 21/1/8 & 21/1/8 & 8/21/1 & 4/10/16 & 3/27/0 & - \\
                \midrule
                \multirow{2}{*}{Ave ranks:} & 16.1 & 17.0 & 16.9 & 12.1 & 7.2 & 20.1 & 17.9 & 10.5 & 10.0 & 7.1 & 18.4 & 13.9 \\
                ~ & 20.2 & 15.6 & 11.3 & 6.4 & 10.9 & 17.9 & 12.1 & 11.8 & 7.9 & \textbf{4.7} & 6.9 & 6.4 \\
			\bottomrule
		\end{tabular}
	}
\end{sidewaystable}

\subsubsection{Results on CEC2022 benchmark functions} \label{sec:4.2.2}
\begin{figure}[!ht]
    \centering
    \includegraphics[width=16cm]{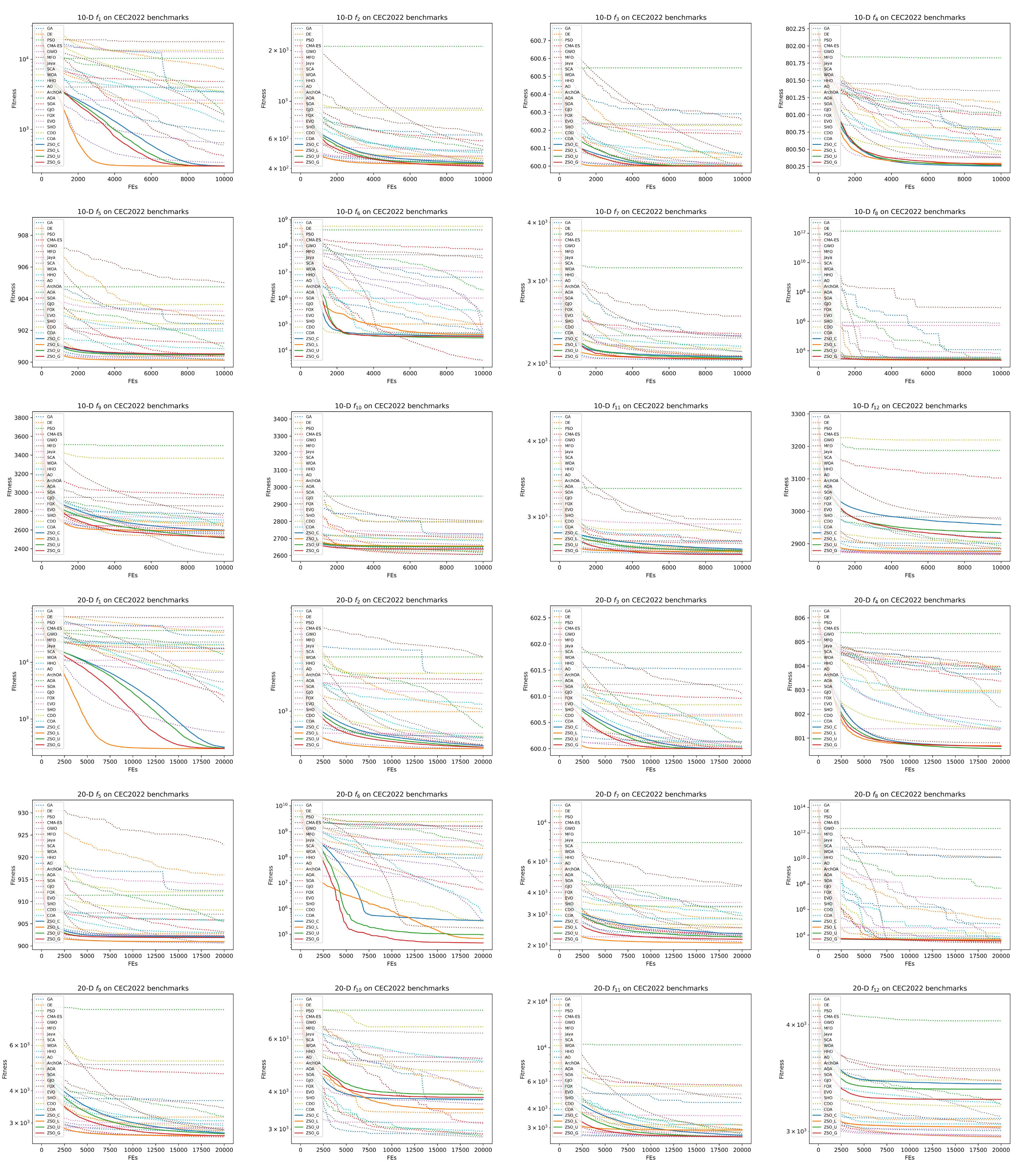}
    \caption{Convergence curves of competitor MAs on CEC2022 benchmark functions.}
    \label{fig:4.2.2.1}
\end{figure}

Figure \ref{fig:4.2.2.1} and presents the convergence curves of competitor algorithms on CEC2022 benchmark functions, Tables \ref{tbl:4.2.2.1} and \ref{tbl:4.2.2.2} summarize the corrected p-value and significance level between ZSO\_Gauss with other competitor algorithms.

\begin{sidewaystable}[htbp]
	\scriptsize
	\centering
	\renewcommand\arraystretch{1.1}
	\caption{Experimental and statistical results between ZSO\_Gauss and other competitor algorithms on 10-D CEC2022 benchmark functions. The mean value and standard deviation (std) of 30 trial runs, the corrected p-value, and the significance level are provided.}
	\label{tbl:4.2.2.1}
	\resizebox{\columnwidth}{!}{
		\begin{tabular}{ccccccccccccccccccccccccc}
			\toprule
			\multicolumn{2}{c}{\multirow{2}{*}{Func.}} & GA & DE & PSO & CMA-ES & GWO & MFO & Jaya & SCA & WOA & HHO & AO & ArchOA \\
                ~ & ~ & AOA & SOA & GJO & FOX & EVO & SHO & CDO & COA & ZSO\_C & ZSO\_L & ZSO\_U & ZSO\_G \\
			\midrule
			\multirow{4}{*}{$f_1$} & \multirow{2}{*}{mean} & 9.350e+02 $+$ & 7.142e+03 $+$ & 1.655e+03 $+$ & 4.216e+02 $+$ & 3.389e+02 $+$ & 1.783e+04 $+$ & 1.260e+04 $+$ & 6.022e+02 $+$ & 3.496e+03 $+$ & 1.284e+03 $+$ & 3.459e+03 $+$ & 2.363e+03 $+$ \\
                ~ & ~ & 1.033e+04 $+$ & 4.831e+03 $+$ & 6.475e+02 $+$ & 1.615e+03 $+$ & 2.612e+03 $+$ & 4.004e+03 $+$ & 1.348e+04 $+$ & 3.362e+03 $+$ & 3.026e+02 $+$ & \textbf{3.000e+02} $-$ & 3.009e+02 $+$ & 3.006e+02 \\
                ~ & \multirow{2}{*}{std} & 3.026e+02 & 1.731e+03 & 1.560e+03 & 4.010e+01 & 3.606e+01 & 5.272e+03 & 3.595e+03 & 1.210e+02 & 1.591e+03 & 6.541e+02 & 1.572e+03 & 1.614e+03 \\
                ~ & ~ & 1.550e+03 & 1.337e+03 & 6.252e+02 & 1.548e+03 & 1.628e+03 & 1.314e+03 & 1.263e+03 & 1.332e+03 & 2.102e+00 & 4.493e-03 & 7.411e-01 & 7.590e-01 \\
                \midrule
			\multirow{4}{*}{$f_2$} & \multirow{2}{*}{mean} & 4.360e+02 $+$ & 4.601e+02 $+$ & 5.159e+02 $+$ & 4.170e+02 $+$ & 4.199e+02 $+$ & 6.396e+02 $+$ & 5.130e+02 $+$ & 4.368e+02 $+$ & 4.580e+02 $+$ & 4.954e+02 $+$ & 6.313e+02 $+$ & 4.741e+02 $+$ \\
                ~ & ~ & 2.115e+03 $+$ & 5.840e+02 $+$ & 4.467e+02 $+$ & 5.427e+02 $+$ & 4.733e+02 $+$ & 9.100e+02 $+$ & 8.864e+02 $+$ & 5.079e+02 $+$ & 4.326e+02 $+$ & 4.258e+02 $+$ & 4.287e+02 $+$ & \textbf{4.136e+02} \\
                ~ & \multirow{2}{*}{std} & 2.853e+01 & 1.426e+01 & 4.474e+01 & 3.039e+00 & 2.199e+01 & 1.263e+02 & 2.145e+01 & 1.399e+01 & 6.122e+01 & 9.147e+01 & 1.077e+02 & 4.004e+01 \\
                ~ & ~ & 8.715e+02 & 7.735e+01 & 1.880e+01 & 8.156e+01 & 3.304e+01 & 2.956e+02 & 3.575e+01 & 4.431e+01 & 3.070e+01 & 3.105e+01 & 3.286e+01 & 2.450e+01 \\
			\multirow{4}{*}{$f_3$} & \multirow{2}{*}{mean} & 6.000e+02 $+$ & 6.001e+02 $+$ & 6.000e+02 $+$ & 6.000e+02 $+$ & 6.000e+02 $+$ & 6.003e+02 $+$ & 6.002e+02 $+$ & 6.000e+02 $+$ & 6.000e+02 $+$ & \textbf{6.000e+02} $-$ & 6.003e+02 $+$ & 6.000e+02 $+$ \\
                ~ & ~ & 6.005e+02 $+$ & 6.002e+02 $+$ & 6.000e+02 $+$ & 6.001e+02 $+$ & 6.000e+02 $+$ & 6.002e+02 $+$ & 6.002e+02 $+$ & 6.001e+02 $+$ & 6.000e+02 $+$ & 6.000e+02 $+$ & 6.000e+02 $+$ & 6.000e+02 \\
                ~ & \multirow{2}{*}{std} & 6.576e-04 & 1.403e-02 & 6.020e-03 & 6.556e-04 & 5.326e-04 & 9.207e-02 & 3.712e-02 & 2.291e-03 & 1.329e-02 & 3.912e-03 & 1.194e-01 & 4.231e-02 \\
                ~ & ~ & 1.321e-01 & 4.186e-02 & 1.166e-02 & 4.769e-02 & 1.275e-02 & 8.384e-02 & 3.743e-03 & 3.811e-02 & 1.484e-05 & 2.156e-07 & 1.685e-05 & 3.199e-06 \\
			\multirow{4}{*}{$f_4$} & \multirow{2}{*}{mean} & 8.008e+02 $+$ & 8.012e+02 $+$ & 8.010e+02 $+$ & 8.010e+02 $+$ & 8.004e+02 $\approx$ & 8.010e+02 $+$ & 8.010e+02 $+$ & 8.005e+02 $+$ & 8.005e+02 $+$ & 8.006e+02 $+$ & 8.008e+02 $+$ & 8.006e+02 $+$ \\
                ~ & ~ & 8.018e+02 $+$ & 8.007e+02 $+$ & 8.004e+02 $+$ & 8.004e+02 $\approx$ & 8.004e+02 $+$ & 8.014e+02 $+$ & 8.008e+02 $+$ & 8.006e+02 $+$ & \textbf{8.003e+02} $\approx$ & 8.003e+02 $\approx$ & 8.003e+02 $\approx$ & 8.003e+02 \\
                ~ & \multirow{2}{*}{std} & 2.537e-01 & 1.955e-01 & 1.917e-01 & 1.567e-01 & 3.302e-01 & 3.229e-01 & 2.217e-01 & 1.408e-01 & 2.690e-01 & 3.890e-01 & 2.547e-01 & 2.674e-01 \\
                ~ & ~ & 4.106e-01 & 1.520e-01 & 1.533e-01 & 2.772e-01 & 1.300e-01 & 3.955e-01 & 2.606e-01 & 1.709e-01 & 1.747e-01 & 2.092e-01 & 1.262e-01 & 1.680e-01 \\
                \midrule
			\multirow{4}{*}{$f_5$} & \multirow{2}{*}{mean} & 9.002e+02 $\approx$ & 9.026e+02 $+$ & 9.010e+02 $+$ & 9.003e+02 $\approx$ & \textbf{9.001e+02} $-$ & 9.050e+02 $+$ & 9.032e+02 $+$ & 9.001e+02 $-$ & 9.024e+02 $+$ & 9.020e+02 $+$ & 9.024e+02 $+$ & 9.004e+02 $\approx$ \\
                ~ & ~ & 9.048e+02 $+$ & 9.012e+02 $+$ & 9.004e+02 $\approx$ & 9.029e+02 $+$ & 9.004e+02 $\approx$ & 9.021e+02 $+$ & 9.036e+02 $+$ & 9.009e+02 $+$ & 9.005e+02 $\approx$ & 9.001e+02 $-$ & 9.005e+02 $\approx$ & 9.005e+02 \\
                ~ & \multirow{2}{*}{std} & 1.825e-01 & 7.560e-01 & 6.036e-01 & 6.385e-02 & 1.854e-01 & 1.155e+00 & 5.043e-01 & 2.946e-02 & 1.829e+00 & 1.505e+00 & 1.209e+00 & 2.582e-01 \\
                ~ & ~ & 1.845e+00 & 2.693e-01 & 2.775e-01 & 1.877e+00 & 2.101e-01 & 1.184e+00 & 7.528e-01 & 3.334e-01 & 8.398e-01 & 2.369e-01 & 5.637e-01 & 6.318e-01 \\
			\multirow{4}{*}{$f_6$} & \multirow{2}{*}{mean} & 5.581e+04 $+$ & 1.001e+05 $+$ & 2.025e+06 $+$ & \textbf{4.096e+03} $-$ & 4.670e+04 $+$ & 3.255e+07 $+$ & 1.015e+07 $+$ & 2.051e+05 $+$ & 3.568e+04 $\approx$ & 4.287e+04 $+$ & 6.139e+06 $+$ & 9.932e+04 $+$ \\
                ~ & ~ & 4.085e+08 $+$ & 7.435e+07 $+$ & 6.164e+04 $+$ & 4.642e+04 $+$ & 9.924e+05 $+$ & 4.289e+07 $+$ & 5.630e+08 $+$ & 3.135e+05 $+$ & 3.527e+04 $\approx$ & 3.977e+04 $\approx$ & 2.964e+04 $\approx$ & 3.354e+04 \\
                ~ & \multirow{2}{*}{std} & 4.197e+04 & 2.880e+04 & 4.474e+06 & 6.088e+02 & 2.313e+04 & 2.907e+07 & 7.350e+06 & 1.008e+05 & 1.517e+04 & 1.385e+04 & 4.834e+06 & 1.494e+05 \\
                ~ & ~ & 3.192e+08 & 7.157e+07 & 3.849e+04 & 1.211e+04 & 1.892e+06 & 3.603e+07 & 4.181e+08 & 4.123e+05 & 1.480e+04 & 1.670e+04 & 1.368e+04 & 1.427e+04 \\
			\multirow{4}{*}{$f_7$} & \multirow{2}{*}{mean} & 2.053e+03 $+$ & 2.072e+03 $+$ & 2.124e+03 $+$ & 2.077e+03 $+$ & 2.049e+03 $\approx$ & 2.522e+03 $+$ & 2.294e+03 $+$ & 2.068e+03 $+$ & 2.157e+03 $+$ & 2.177e+03 $+$ & 2.291e+03 $+$ & 2.048e+03 $+$ \\
                ~ & ~ & 3.204e+03 $+$ & 2.313e+03 $+$ & 2.124e+03 $+$ & 2.283e+03 $+$ & 2.047e+03 $\approx$ & 2.268e+03 $+$ & 3.836e+03 $+$ & 2.066e+03 $+$ & 2.069e+03 $+$ & 2.048e+03 $\approx$ & 2.054e+03 $+$ & \textbf{2.043e+03} \\
                ~ & \multirow{2}{*}{std} & 5.519e+00 & 1.017e+01 & 5.245e+01 & 1.483e+01 & 2.645e+01 & 2.111e+02 & 6.430e+01 & 1.358e+01 & 9.381e+01 & 1.057e+02 & 1.310e+02 & 1.448e+01 \\
                ~ & ~ & 7.335e+02 & 1.406e+02 & 9.190e+01 & 1.391e+02 & 2.412e+01 & 1.195e+02 & 1.169e+03 & 3.462e+01 & 4.196e+01 & 3.962e+01 & 2.442e+01 & 1.669e+01 \\
			\multirow{4}{*}{$f_8$} & \multirow{2}{*}{mean} & 2.577e+03 $+$ & 2.257e+03 $\approx$ & 2.502e+03 $+$ & \textbf{2.229e+03} $-$ & 2.743e+03 $+$ & 8.646e+06 $+$ & 6.430e+03 $+$ & 2.314e+03 $+$ & 3.312e+03 $+$ & 3.207e+03 $+$ & 1.171e+04 $+$ & 2.336e+03 $+$ \\
                ~ & ~ & 1.400e+12 $+$ & 2.297e+03 $+$ & 3.697e+03 $+$ & 2.705e+03 $+$ & 5.265e+05 $+$ & 7.710e+05 $+$ & 2.893e+03 $+$ & 2.339e+03 $+$ & 2.386e+03 $+$ & 2.321e+03 $\approx$ & 2.348e+03 $+$ & 2.279e+03 \\
                ~ & \multirow{2}{*}{std} & 4.167e+02 & 1.029e+01 & 4.845e+02 & 1.936e+00 & 6.502e+02 & 3.449e+07 & 1.822e+03 & 5.039e+01 & 9.562e+02 & 9.692e+02 & 4.065e+04 & 1.439e+02 \\
                ~ & ~ & 4.354e+12 & 9.298e+01 & 6.211e+02 & 4.524e+02 & 1.663e+06 & 2.111e+06 & 3.718e+02 & 9.352e+01 & 1.637e+02 & 1.988e+02 & 2.033e+02 & 7.708e+01 \\
                \midrule
			\multirow{4}{*}{$f_9$} & \multirow{2}{*}{mean} & 2.580e+03 $+$ & 2.670e+03 $+$ & 2.630e+03 $+$ & 2.602e+03 $+$ & 2.566e+03 $\approx$ & 2.850e+03 $+$ & 2.736e+03 $+$ & \textbf{2.338e+03} $\approx$ & 2.657e+03 $+$ & 2.707e+03 $+$ & 2.777e+03 $+$ & 2.539e+03 $\approx$ \\
                ~ & ~ & 3.501e+03 $+$ & 2.972e+03 $+$ & 2.555e+03 $\approx$ & 2.757e+03 $+$ & 2.647e+03 $+$ & 2.945e+03 $+$ & 3.366e+03 $+$ & 2.684e+03 $+$ & 2.598e+03 $+$ & 2.595e+03 $+$ & 2.516e+03 $\approx$ & 2.525e+03 \\
                ~ & \multirow{2}{*}{std} & 1.620e+02 & 2.669e+01 & 1.477e+02 & 1.106e+02 & 1.709e+02 & 6.963e+01 & 1.160e+01 & 1.268e+01 & 2.015e+02 & 1.659e+02 & 7.140e+01 & 1.615e+02 \\
                ~ & ~ & 3.106e+02 & 1.399e+02 & 1.927e+02 & 1.753e+02 & 1.731e+02 & 1.697e+02 & 9.523e+01 & 8.943e+01 & 2.098e+02 & 1.809e+02 & 2.069e+02 & 2.087e+02 \\
			\multirow{4}{*}{$f_{10}$} & \multirow{2}{*}{mean} & 2.645e+03 $\approx$ & 2.624e+03 $\approx$ & 2.616e+03 $\approx$ & \textbf{2.606e+03} $\approx$ & 2.643e+03 $\approx$ & 2.795e+03 $+$ & 2.725e+03 $+$ & 2.606e+03 $\approx$ & 2.694e+03 $+$ & 2.687e+03 $+$ & 2.726e+03 $+$ & 2.625e+03 $\approx$ \\
                ~ & ~ & 2.948e+03 $+$ & 2.705e+03 $+$ & 2.664e+03 $+$ & 2.803e+03 $+$ & 2.653e+03 $+$ & 2.715e+03 $+$ & 2.800e+03 $+$ & 2.628e+03 $\approx$ & 2.653e+03 $+$ & 2.656e+03 $+$ & 2.652e+03 $+$ & 2.638e+03 \\
                ~ & \multirow{2}{*}{std} & 6.447e+01 & 5.490e+00 & 8.243e+00 & 1.069e+00 & 5.990e+01 & 9.796e+01 & 1.042e+02 & 3.173e+00 & 1.015e+02 & 8.678e+01 & 1.049e+02 & 3.243e+01 \\
                ~ & ~ & 2.581e+02 & 5.758e+01 & 7.102e+01 & 2.827e+02 & 6.024e+01 & 9.403e+01 & 3.140e+01 & 1.169e+01 & 6.862e+01 & 6.932e+01 & 7.247e+01 & 6.315e+01 \\
			\multirow{4}{*}{$f_{11}$} & \multirow{2}{*}{mean} & 2.608e+03 $+$ & 2.659e+03 $+$ & 2.628e+03 $+$ & 2.742e+03 $+$ & 2.632e+03 $+$ & 2.969e+03 $+$ & 2.919e+03 $+$ & 2.608e+03 $+$ & 2.622e+03 $+$ & 2.629e+03 $+$ & 2.731e+03 $+$ & 2.650e+03 $+$ \\
                ~ & ~ & 3.340e+03 $+$ & 2.711e+03 $+$ & 2.612e+03 $+$ & 2.820e+03 $+$ & 2.659e+03 $+$ & 2.739e+03 $+$ & 2.854e+03 $+$ & 2.649e+03 $+$ & 2.654e+03 $+$ & 2.633e+03 $+$ & 2.639e+03 $+$ & \textbf{2.601e+03} \\
                ~ & \multirow{2}{*}{std} & 6.685e+00 & 1.517e+02 & 2.065e+01 & 3.007e+02 & 1.568e+02 & 2.398e+02 & 3.508e+02 & 2.743e+00 & 2.856e+01 & 2.755e+01 & 2.053e+02 & 4.782e+01 \\
                ~ & ~ & 2.632e+02 & 5.555e+01 & 1.377e+01 & 3.316e+02 & 1.571e+02 & 1.614e+02 & 2.699e+01 & 1.773e+01 & 1.751e+02 & 1.571e+02 & 1.872e+02 & 3.614e+00 \\
			\multirow{4}{*}{$f_{12}$} & \multirow{2}{*}{mean} & 2.870e+03 $-$ & 2.868e+03 $-$ & 2.895e+03 $-$ & \textbf{2.867e+03} $-$ & 2.867e+03 $-$ & 2.884e+03 $-$ & 2.871e+03 $-$ & 2.887e+03 $-$ & 2.907e+03 $-$ & 2.919e+03 $\approx$ & 2.901e+03 $-$ & 2.887e+03 $-$ \\
                ~ & ~ & 3.188e+03 $+$ & 3.103e+03 $+$ & 2.870e+03 $-$ & 2.976e+03 $+$ & 2.875e+03 $-$ & 2.981e+03 $+$ & 3.220e+03 $+$ & 2.878e+03 $-$ & 2.958e+03 $+$ & 2.877e+03 $-$ & 2.934e+03 $+$ & 2.916e+03 \\
                ~ & \multirow{2}{*}{std} & 1.700e+00 & 6.642e-01 & 1.032e+01 & 5.348e-01 & 2.193e+00 & 8.925e+00 & 9.600e-01 & 4.228e+00 & 5.119e+01 & 4.758e+01 & 3.461e+01 & 2.728e+01 \\
                ~ & ~ & 1.064e+02 & 6.205e+01 & 7.433e+00 & 4.861e+01 & 6.371e+00 & 5.973e+01 & 1.075e+02 & 4.481e+00 & 4.531e+01 & 1.243e+01 & 3.412e+01 & 3.313e+01 \\
            \midrule
                \multicolumn{2}{c}{\multirow{2}{*}{$+$/$\approx$/$-$:}} & 9/2/1 & 9/2/1 & 10/1/1 & 7/2/3 & 6/4/2 & 11/0/1 & 11/0/1 & 8/2/2 & 10/1/1 & 10/1/1 & 11/0/1 & 8/3/1 \\
                ~ & ~ & 12/0/0 & 12/0/0 & 9/2/1 & 11/1/0 & 9/2/1 & 12/0/0 & 12/0/0 & 10/1/1 & 9/3/0 & 5/4/3 & 8/4/0 & - \\
            \midrule
                \multicolumn{2}{c}{\multirow{2}{*}{Ave ranks:}} & 7.2 & 12.4 & 12.1 & 7.0 & 6.0 & 20.9 & 18.1 & 7.0 & 12.4 & 12.6 & 18.3 & 9.6 \\
                ~ & ~ & 23.4 & 17.3 & 9.5 & 15.9 & 11.1 & 19.9 & 21.4 & 12.0 & 8.4 & 5.5 & 6.6 & \textbf{4.5} \\
		  \bottomrule
		\end{tabular}
	}
\end{sidewaystable}

\begin{sidewaystable}[htbp]
	\scriptsize
	\centering
	\renewcommand\arraystretch{1.1}
	\caption{Experimental and statistical results between ZSO\_Gauss and other competitor algorithms on 20-D CEC2022 benchmark functions.}
	\label{tbl:4.2.2.2}
	\resizebox{\columnwidth}{!}{
		\begin{tabular}{ccccccccccccccccccccccccc}
			\toprule
			\multicolumn{2}{c}{\multirow{2}{*}{Func.}} & GA & DE & PSO & CMA-ES & GWO & MFO & Jaya & SCA & WOA & HHO & AO & ArchOA \\
                ~ & ~ & AOA & SOA & GJO & FOX & EVO & SHO & CDO & COA & ZSO\_C & ZSO\_L & ZSO\_U & ZSO\_G \\
			\midrule
			\multirow{4}{*}{$f_1$} & \multirow{2}{*}{mean} & 1.776e+04 $+$ & 3.379e+04 $+$ & 1.356e+04 $+$ & 2.546e+03 $+$ & 5.893e+02 $+$ & 6.146e+04 $+$ & 4.188e+04 $+$ & 4.506e+03 $+$ & 6.834e+03 $+$ & 3.252e+03 $+$ & 2.993e+04 $+$ & 1.553e+04 $+$ \\
                ~ & ~ & 3.615e+04 $+$ & 1.733e+04 $+$ & 6.642e+03 $+$ & 1.945e+03 $+$ & 1.084e+04 $+$ & 2.272e+04 $+$ & 1.995e+04 $+$ & 2.063e+04 $+$ & 3.195e+02 $+$ & \textbf{3.000e+02} $-$ & 3.075e+02 $+$ & 3.036e+02 \\
                ~ & \multirow{2}{*}{std} & 2.479e+03 & 5.789e+03 & 5.981e+03 & 4.326e+02 & 2.175e+02 & 1.063e+04 & 8.482e+03 & 1.144e+03 & 2.906e+03 & 1.301e+03 & 8.785e+03 & 4.349e+03 \\
                ~ & ~ & 7.961e+03 & 2.822e+03 & 2.059e+03 & 2.026e+03 & 5.998e+03 & 5.116e+03 & 1.864e+02 & 4.377e+03 & 2.667e+01 & 4.047e-03 & 9.440e+00 & 3.298e+00 \\
                \midrule
			\multirow{4}{*}{$f_2$} & \multirow{2}{*}{mean} & 5.956e+02 $+$ & 1.047e+03 $+$ & 7.142e+02 $+$ & 5.099e+02 $+$ & 4.708e+02 $\approx$ & 2.961e+03 $+$ & 1.436e+03 $+$ & 5.513e+02 $+$ & 6.029e+02 $+$ & 5.858e+02 $+$ & 2.137e+03 $+$ & 9.896e+02 $+$ \\
                ~ & ~ & 2.980e+03 $+$ & 1.886e+03 $+$ & 5.818e+02 $+$ & 5.028e+02 $+$ & 6.419e+02 $+$ & 1.748e+03 $+$ & 2.142e+03 $+$ & 1.137e+03 $+$ & 4.994e+02 $+$ & \textbf{4.697e+02} $-$ & 4.843e+02 $\approx$ & 4.834e+02 \\
                ~ & \multirow{2}{*}{std} & 2.987e+01 & 1.488e+02 & 1.992e+02 & 1.540e+01 & 1.736e+01 & 9.742e+02 & 1.747e+02 & 2.683e+01 & 7.697e+01 & 5.248e+01 & 5.072e+02 & 2.433e+02 \\
                ~ & ~ & 6.602e+02 & 3.988e+02 & 4.948e+01 & 2.351e+01 & 6.776e+01 & 4.843e+02 & 4.483e+01 & 1.811e+02 & 3.211e+01 & 2.297e+01 & 1.995e+01 & 2.680e+01 \\
			\multirow{4}{*}{$f_3$} & \multirow{2}{*}{mean} & 6.001e+02 $+$ & 6.004e+02 $+$ & 6.001e+02 $+$ & 6.000e+02 $+$ & 6.000e+02 $+$ & 6.010e+02 $+$ & 6.007e+02 $+$ & 6.001e+02 $+$ & 6.001e+02 $+$ & 6.000e+02 $+$ & 6.015e+02 $+$ & 6.006e+02 $+$ \\
                ~ & ~ & 6.018e+02 $+$ & 6.010e+02 $+$ & 6.001e+02 $+$ & 6.000e+02 $+$ & 6.001e+02 $+$ & 6.012e+02 $+$ & 6.008e+02 $+$ & 6.005e+02 $+$ & 6.000e+02 $+$ & \textbf{6.000e+02} $-$ & 6.000e+02 $+$ & 6.000e+02 \\
                ~ & \multirow{2}{*}{std} & 1.988e-02 & 5.529e-02 & 6.535e-02 & 7.244e-03 & 6.702e-03 & 2.730e-01 & 7.189e-02 & 1.049e-02 & 3.417e-02 & 2.556e-02 & 4.205e-01 & 2.823e-01 \\
                ~ & ~ & 1.422e-01 & 8.123e-02 & 7.141e-02 & 2.062e-03 & 3.719e-02 & 2.525e-01 & 1.061e-02 & 1.316e-01 & 1.621e-04 & 1.559e-07 & 4.524e-05 & 3.190e-05 \\
			\multirow{4}{*}{$f_4$} & \multirow{2}{*}{mean} & 8.039e+02 $+$ & 8.040e+02 $+$ & 8.039e+02 $+$ & 8.040e+02 $+$ & 8.013e+02 $+$ & 8.037e+02 $+$ & 8.039e+02 $+$ & 8.023e+02 $+$ & 8.014e+02 $+$ & 8.014e+02 $+$ & 8.037e+02 $+$ & 8.029e+02 $+$ \\
                ~ & ~ & 8.053e+02 $+$ & 8.034e+02 $+$ & 8.017e+02 $+$ & 8.008e+02 $\approx$ & 8.014e+02 $+$ & 8.046e+02 $+$ & 8.030e+02 $+$ & 8.029e+02 $+$ & 8.007e+02 $\approx$ & 8.007e+02 $\approx$ & \textbf{8.006e+02} $\approx$ & 8.007e+02 \\
                ~ & \multirow{2}{*}{std} & 3.693e-01 & 4.064e-01 & 4.039e-01 & 3.567e-01 & 1.317e+00 & 6.759e-01 & 4.248e-01 & 3.094e-01 & 6.231e-01 & 5.911e-01 & 5.716e-01 & 8.151e-01 \\
                ~ & ~ & 5.978e-01 & 4.693e-01 & 6.451e-01 & 3.403e-01 & 5.420e-01 & 6.141e-01 & 5.442e-01 & 4.482e-01 & 3.435e-01 & 3.702e-01 & 2.662e-01 & 3.410e-01 \\
                \midrule
			\multirow{4}{*}{$f_5$} & \multirow{2}{*}{mean} & 9.022e+02 $+$ & 9.160e+02 $+$ & 9.054e+02 $+$ & 9.032e+02 $+$ & 9.007e+02 $-$ & 9.228e+02 $+$ & 9.139e+02 $+$ & \textbf{9.007e+02} $-$ & 9.081e+02 $+$ & 9.030e+02 $+$ & 9.125e+02 $+$ & 9.028e+02 $+$ \\
                ~ & ~ & 9.114e+02 $+$ & 9.058e+02 $+$ & 9.020e+02 $\approx$ & 9.024e+02 $+$ & 9.024e+02 $+$ & 9.072e+02 $+$ & 9.122e+02 $+$ & 9.058e+02 $+$ & 9.021e+02 $\approx$ & 9.010e+02 $-$ & 9.019e+02 $\approx$ & 9.019e+02 \\
                ~ & \multirow{2}{*}{std} & 4.300e-01 & 2.303e+00 & 3.328e+00 & 4.787e-01 & 6.336e-01 & 4.536e+00 & 1.521e+00 & 2.614e-01 & 2.867e+00 & 1.705e+00 & 2.989e+00 & 1.067e+00 \\
                ~ & ~ & 3.157e+00 & 1.054e+00 & 9.226e-01 & 7.808e-01 & 1.036e+00 & 2.780e+00 & 2.363e+00 & 1.522e+00 & 8.752e-01 & 6.906e-01 & 7.381e-01 & 1.068e+00 \\
			\multirow{4}{*}{$f_6$} & \multirow{2}{*}{mean} & 9.296e+07 $+$ & 2.221e+08 $+$ & 2.708e+08 $+$ & 5.283e+06 $+$ & 3.501e+05 $+$ & 7.551e+08 $+$ & 4.289e+08 $+$ & 1.008e+07 $+$ & 3.077e+05 $+$ & 6.808e+05 $+$ & 1.649e+09 $+$ & 1.279e+08 $+$ \\
                ~ & ~ & 4.492e+09 $+$ & 1.591e+09 $+$ & 2.760e+07 $+$ & 1.750e+05 $+$ & 1.756e+07 $+$ & 1.311e+09 $+$ & 2.433e+09 $+$ & 1.110e+08 $+$ & 3.444e+05 $+$ & 6.894e+04 $+$ & 9.565e+04 $+$ & \textbf{4.574e+04} \\
                ~ & \multirow{2}{*}{std} & 3.584e+07 & 8.545e+07 & 2.109e+08 & 2.076e+06 & 9.275e+05 & 8.106e+08 & 2.105e+08 & 5.353e+06 & 8.818e+05 & 1.726e+06 & 1.306e+09 & 1.697e+08 \\
                ~ & ~ & 1.864e+09 & 8.326e+08 & 4.131e+07 & 4.090e+04 & 3.359e+07 & 6.561e+08 & 4.033e+07 & 6.975e+07 & 7.947e+04 & 2.411e+04 & 2.459e+04 & 1.608e+04 \\
			\multirow{4}{*}{$f_7$} & \multirow{2}{*}{mean} & 2.295e+03 $+$ & 2.823e+03 $+$ & 2.958e+03 $+$ & 2.286e+03 $+$ & 2.101e+03 $-$ & 4.390e+03 $+$ & 3.513e+03 $+$ & 2.228e+03 $+$ & 3.303e+03 $+$ & 3.048e+03 $+$ & 3.335e+03 $+$ & 2.518e+03 $+$ \\
                ~ & ~ & 7.693e+03 $+$ & 2.350e+03 $+$ & 2.263e+03 $\approx$ & 2.522e+03 $+$ & 2.240e+03 $\approx$ & 4.345e+03 $+$ & 2.633e+03 $+$ & 2.821e+03 $+$ & 2.333e+03 $+$ & \textbf{2.067e+03} $-$ & 2.237e+03 $\approx$ & 2.168e+03 \\
                ~ & \multirow{2}{*}{std} & 7.169e+01 & 2.064e+02 & 4.135e+02 & 7.372e+01 & 5.380e+01 & 8.572e+02 & 4.029e+02 & 5.886e+01 & 8.203e+02 & 4.861e+02 & 6.720e+02 & 2.680e+02 \\
                ~ & ~ & 1.694e+03 & 8.694e+01 & 2.309e+02 & 3.262e+02 & 1.684e+02 & 9.914e+02 & 1.143e+02 & 2.775e+02 & 2.044e+02 & 2.957e+01 & 1.593e+02 & 9.204e+01 \\
			\multirow{4}{*}{$f_8$} & \multirow{2}{*}{mean} & 6.602e+04 $+$ & 1.773e+05 $+$ & 4.653e+07 $+$ & \textbf{2.412e+03} $-$ & 5.254e+03 $+$ & 1.318e+10 $+$ & 7.257e+06 $+$ & 3.812e+03 $+$ & 5.679e+03 $+$ & 5.387e+03 $+$ & 1.257e+10 $+$ & 4.699e+03 $+$ \\
                ~ & ~ & 2.287e+12 $+$ & 3.542e+03 $+$ & 7.699e+03 $+$ & 2.748e+03 $-$ & 3.832e+04 $+$ & 5.075e+10 $+$ & 1.441e+04 $+$ & 7.375e+03 $+$ & 3.473e+03 $\approx$ & 4.211e+03 $+$ & 3.403e+03 $\approx$ & 3.446e+03 \\
                ~ & \multirow{2}{*}{std} & 1.426e+05 & 3.467e+05 & 1.941e+08 & 5.477e+01 & 1.110e+03 & 2.580e+10 & 2.119e+07 & 8.059e+02 & 1.897e+03 & 1.393e+03 & 2.449e+10 & 1.180e+03 \\
                ~ & ~ & 7.645e+12 & 2.829e+02 & 3.292e+03 & 3.471e+02 & 1.686e+05 & 2.434e+11 & 2.021e+04 & 3.850e+03 & 6.674e+02 & 9.289e+02 & 8.062e+02 & 9.676e+02 \\
                \midrule
			\multirow{4}{*}{$f_9$} & \multirow{2}{*}{mean} & 2.791e+03 $+$ & 2.806e+03 $+$ & 3.176e+03 $+$ & \textbf{2.649e+03} $-$ & 2.675e+03 $\approx$ & 3.415e+03 $+$ & 2.984e+03 $+$ & 2.836e+03 $+$ & 2.871e+03 $+$ & 2.932e+03 $+$ & 3.661e+03 $+$ & 3.160e+03 $+$ \\
                ~ & ~ & 8.244e+03 $+$ & 4.657e+03 $+$ & 2.843e+03 $+$ & 2.825e+03 $+$ & 2.880e+03 $+$ & 5.044e+03 $+$ & 5.214e+03 $+$ & 3.058e+03 $+$ & 2.744e+03 $+$ & 2.681e+03 $+$ & 2.726e+03 $+$ & 2.675e+03 \\
                ~ & \multirow{2}{*}{std} & 3.181e+01 & 4.116e+01 & 2.214e+02 & 3.923e+00 & 2.696e+01 & 3.534e+02 & 8.838e+01 & 4.470e+01 & 1.879e+02 & 2.099e+02 & 4.502e+02 & 2.963e+02 \\
                ~ & ~ & 1.891e+03 & 4.118e+02 & 9.513e+01 & 1.098e+02 & 1.078e+02 & 8.332e+02 & 1.483e+02 & 1.665e+02 & 1.030e+02 & 3.388e+01 & 8.379e+01 & 9.994e+01 \\
			\multirow{4}{*}{$f_{10}$} & \multirow{2}{*}{mean} & 2.891e+03 $\approx$ & 4.018e+03 $\approx$ & 2.845e+03 $\approx$ & 3.135e+03 $\approx$ & 4.108e+03 $\approx$ & 5.121e+03 $+$ & 5.174e+03 $+$ & \textbf{2.825e+03} $-$ & 4.672e+03 $+$ & 5.050e+03 $+$ & 3.745e+03 $\approx$ & 3.398e+03 $\approx$ \\
                ~ & ~ & 7.482e+03 $+$ & 2.889e+03 $\approx$ & 3.640e+03 $\approx$ & 5.023e+03 $+$ & 3.163e+03 $\approx$ & 6.274e+03 $+$ & 6.583e+03 $+$ & 2.953e+03 $\approx$ & 3.774e+03 $\approx$ & 3.494e+03 $\approx$ & 3.914e+03 $\approx$ & 3.826e+03 \\
                ~ & \multirow{2}{*}{std} & 1.220e+02 & 1.655e+03 & 3.896e+01 & 9.325e+02 & 1.539e+03 & 1.357e+03 & 1.871e+03 & 7.987e+01 & 1.161e+03 & 1.124e+03 & 1.461e+03 & 1.117e+03 \\
                ~ & ~ & 1.037e+03 & 5.218e+01 & 1.308e+03 & 8.288e+02 & 6.843e+02 & 1.456e+03 & 1.175e+03 & 1.016e+02 & 1.005e+03 & 8.887e+02 & 1.053e+03 & 1.108e+03 \\
			\multirow{4}{*}{$f_{11}$} & \multirow{2}{*}{mean} & 2.652e+03 $+$ & 2.808e+03 $+$ & 2.734e+03 $+$ & 2.620e+03 $+$ & \textbf{2.612e+03} $-$ & 4.668e+03 $+$ & 3.593e+03 $+$ & 2.628e+03 $+$ & 2.939e+03 $+$ & 2.797e+03 $+$ & 4.381e+03 $+$ & 3.134e+03 $+$ \\
                ~ & ~ & 1.044e+04 $+$ & 5.697e+03 $+$ & 2.649e+03 $+$ & 2.922e+03 $+$ & 2.809e+03 $+$ & 4.976e+03 $+$ & 5.598e+03 $+$ & 3.121e+03 $+$ & 2.680e+03 $+$ & 2.896e+03 $+$ & 2.617e+03 $+$ & 2.613e+03 \\
                ~ & \multirow{2}{*}{std} & 7.180e+00 & 3.126e+02 & 1.550e+02 & 3.111e+00 & 6.916e+00 & 4.056e+02 & 6.891e+02 & 4.930e+00 & 5.392e+02 & 3.547e+02 & 1.511e+03 & 4.395e+02 \\
                ~ & ~ & 2.263e+03 & 1.294e+03 & 3.247e+01 & 5.200e+02 & 3.797e+02 & 1.404e+03 & 1.031e+02 & 2.800e+02 & 1.722e+02 & 5.779e+02 & 3.464e+01 & 5.263e+01 \\
			\multirow{4}{*}{$f_{12}$} & \multirow{2}{*}{mean} & 3.006e+03 $-$ & \textbf{2.955e+03} $-$ & 3.225e+03 $\approx$ & 2.957e+03 $-$ & 2.965e+03 $-$ & 3.064e+03 $-$ & 2.971e+03 $-$ & 3.125e+03 $-$ & 3.209e+03 $\approx$ & 3.237e+03 $\approx$ & 3.106e+03 $-$ & 3.095e+03 $-$ \\
                ~ & ~ & 4.042e+03 $+$ & 3.438e+03 $+$ & 3.026e+03 $-$ & 3.547e+03 $+$ & 3.020e+03 $-$ & 3.534e+03 $+$ & 3.446e+03 $+$ & 3.055e+03 $-$ & 3.410e+03 $+$ & 3.037e+03 $-$ & 3.362e+03 $+$ & 3.270e+03 \\
                ~ & \multirow{2}{*}{std} & 1.013e+01 & 4.038e+00 & 6.974e+01 & 7.635e+00 & 1.184e+01 & 5.319e+01 & 8.887e+00 & 2.699e+01 & 1.304e+02 & 1.516e+02 & 7.312e+01 & 7.243e+01 \\
                ~ & ~ & 3.005e+02 & 6.494e+01 & 4.124e+01 & 1.855e+02 & 3.620e+01 & 1.717e+02 & 3.665e+01 & 3.898e+01 & 1.041e+02 & 5.145e+01 & 1.109e+02 & 1.109e+02 \\
            \midrule
                \multicolumn{2}{c}{\multirow{2}{*}{$+$/$\approx$/$-$:}} & 10/1/1 & 10/1/1 & 5/3/4 & 8/1/3 & 10/2/0 & 11/0/1 & 11/0/1 & 9/0/3 & 12/0/0 & 11/1/0 & 10/1/1 & 10/1/1 \\
                ~ & ~ & 12/0/0 & 11/1/0 & 8/3/1 & 10/1/1 & 9/2/1 & 12/0/0 & 12/0/0 & 10/1/1 & 8/4/0 & 4/2/6 & 6/6/0 & - \\
            \midrule
                \multicolumn{2}{c}{\multirow{2}{*}{Ave ranks:}} & 10.7 & 15.1 & 14.0 & 7.1 & 5.1 & 20.1 & 18.0 & 7.5 & 12.6 & 11.9 & 18.8 & 13.2 \\
                ~ & ~ & 23.4 & 15.9 & 9.5 & 9.5 & 10.4 & 21.0 & 19.4 & 13.8 & 7.3 & \textbf{4.5} & 5.5 & 4.7 \\
		  \bottomrule
		\end{tabular}
	}
\end{sidewaystable}

\subsubsection{Results on six engineering problems} \label{sec:4.2.3}
\begin{figure}[!ht]
    \centering
    \includegraphics[width=16cm]{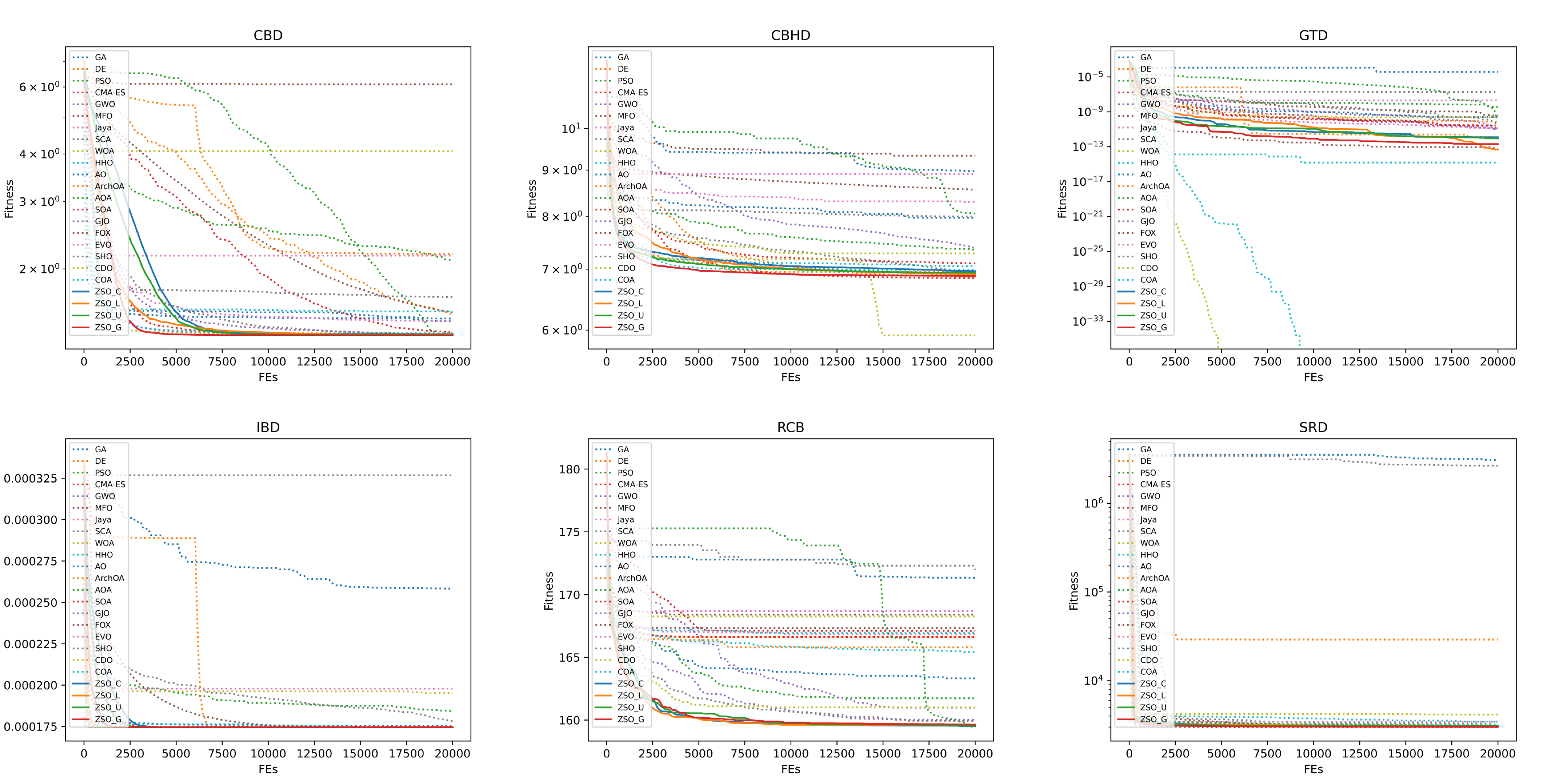}
    \caption{Convergence curves of competitor algorithms on engineering problems.}
    \label{fig:4.2.3.1}
\end{figure}

Figure \ref{fig:4.2.3.1} visualizes the convergence curves of 20 popular MAs when applied to six engineering problems, Table \ref{tbl:4.2.3.1} summarizes the corrected p-value and significance level between ZSO\_Gauss with other competitor algorithms.

\begin{sidewaystable}[htbp]
	\scriptsize
	\centering
	\renewcommand\arraystretch{1.5}
	\caption{Experimental and statistical results between ZSO\_Gauss and other competitor algorithms on engineering problems.}
	\label{tbl:4.2.3.1}
	\resizebox{\columnwidth}{!}{
		\begin{tabular}{ccccccccccccccccccccccccc}
			\toprule
			\multicolumn{2}{c}{\multirow{2}{*}{Func.}} & GA & DE & PSO & CMA-ES & GWO & MFO & Jaya & SCA & WOA & HHO & AO & ArchOA \\
                ~ & ~ & AOA & SOA & GJO & FOX & EVO & SHO & CDO & COA & ZSO\_C & ZSO\_L & ZSO\_U & ZSO\_G \\
			\midrule
			\multirow{4}{*}{CBD} & \multirow{2}{*}{mean} & 1.459e+00 $+$ & 1.520e+00 $+$ & 2.103e+00 $+$ & 1.362e+00 $+$ & \textbf{1.340e+00} $\approx$ & 6.099e+00 $+$ & 1.460e+00 $+$ & 1.345e+00 $+$ & 4.070e+00 $+$ & 1.342e+00 $+$ & 1.482e+00 $+$ & 2.189e+00 $+$ \\
                ~ & ~ & 1.348e+00 $+$ & 1.346e+00 $+$ & 1.340e+00 $\approx$ & 1.537e+00 $+$ & 2.167e+00 $+$ & 1.689e+00 $+$ & 1.358e+00 $+$ & 1.547e+00 $+$ & 1.349e+00 $+$ & 1.340e+00 $\approx$ & 1.343e+00 $+$ & 1.340e+00 \\
                ~ & \multirow{2}{*}{std} & 1.137e-01 & 7.315e-02 & 2.892e-01 & 1.306e-02 & 6.244e-05 & 1.371e+00 & 3.558e-02 & 2.185e-03 & 1.497e+00 & 1.921e-03 & 6.279e-02 & 5.524e-01 \\
                ~ & ~ & 5.317e-03 & 4.505e-03 & 3.264e-04 & 4.388e-01 & 4.522e-01 & 2.421e-01 & 6.720e-03 & 1.634e-01 & 3.598e-03 & 1.219e-04 & 1.642e-03 & 1.554e-04 \\
                \midrule
			\multirow{4}{*}{CBHD} & \multirow{2}{*}{mean} & 7.967e+00 $+$ & 6.865e+00 $\approx$ & 7.360e+00 $+$ & 6.847e+00 $-$ & 6.859e+00 $\approx$ & 9.335e+00 $+$ & 8.300e+00 $+$ & 6.922e+00 $+$ & 7.289e+00 $+$ & 6.990e+00 $\approx$ & 8.976e+00 $+$ & 6.953e+00 $+$ \\
                ~ & ~ & 8.065e+00 $+$ & 7.105e+00 $+$ & 7.396e+00 $+$ & 8.565e+00 $+$ & 8.914e+00 $+$ & 8.001e+00 $+$ & \textbf{5.923e+00} $-$ & 7.056e+00 $+$ & 6.966e+00 $+$ & 6.912e+00 $+$ & 6.938e+00 $+$ & 6.885e+00 \\
                ~ & \multirow{2}{*}{std} & 7.019e-01 & 6.617e-03 & 2.349e-01 & 1.567e-03 & 1.034e-02 & 7.434e-01 & 5.241e-01 & 2.733e-02 & 3.598e-01 & 1.310e-01 & 1.155e+00 & 5.856e-02 \\
                ~ & ~ & 1.366e+00 & 7.687e-02 & 5.028e-01 & 1.040e+00 & 8.294e-01 & 6.612e-01 & 2.037e+00 & 1.129e-01 & 6.649e-02 & 5.454e-02 & 6.186e-02 & 3.413e-02 \\
                \midrule
			\multirow{4}{*}{GTD} & \multirow{2}{*}{mean} & 2.487e-10 $+$ & 7.742e-11 $+$ & 3.690e-09 $+$ & 1.404e-11 $+$ & 9.970e-12 $+$ & 3.697e-10 $+$ & 1.288e-11 $+$ & 4.815e-11 $+$ & \textbf{0.000e+00} $-$ & \textbf{0.000e+00} $-$ & 3.622e-05 $+$ & 1.030e-12 $+$ \\
                ~ & ~ & 2.527e-10 $+$ & 2.726e-11 $+$ & 7.404e-12 $+$ & 6.663e-14 $-$ & 2.089e-08 $+$ & 1.904e-07 $+$ & 2.336e-10 $+$ & 1.558e-15 $-$ & 1.231e-12 $+$ & 4.744e-14 $-$ & 8.838e-13 $+$ & 1.990e-13 \\
                ~ & \multirow{2}{*}{std} & 5.445e-10 & 1.449e-10 & 8.083e-09 & 1.625e-11 & 2.111e-11 & 1.073e-09 & 3.067e-11 & 1.110e-10 & 0.000e+00 & 0.000e+00 & 1.940e-04 & 1.996e-12 \\
                ~ & ~ & 5.275e-10 & 8.577e-11 & 9.276e-12 & 1.620e-13 & 8.880e-08 & 5.944e-07 & 2.907e-10 & 8.388e-15 & 1.514e-12 & 9.503e-14 & 2.784e-12 & 3.292e-13 \\
                \midrule
			\multirow{4}{*}{IBD} & \multirow{2}{*}{mean} & 1.753e-04 $+$ & 1.746e-04 $+$ & 1.840e-04 $+$ & 1.746e-04 $-$ & 1.746e-04 $\approx$ & 1.748e-04 $+$ & 1.746e-04 $\approx$ & 1.784e-04 $+$ & 1.952e-04 $+$ & \textbf{1.746e-04} $-$ & 2.582e-04 $+$ & 1.748e-04 $+$ \\
                ~ & ~ & 1.746e-04 $+$ & 1.746e-04 $-$ & 1.746e-04 $\approx$ & 1.746e-04 $\approx$ & 1.979e-04 $+$ & 3.267e-04 $+$ & 1.746e-04 $+$ & 1.746e-04 $-$ & 1.746e-04 $\approx$ & 1.746e-04 $\approx$ & 1.746e-04 $\approx$ & 1.746e-04 \\
                ~ & \multirow{2}{*}{std} & 9.118e-07 & 2.754e-09 & 3.729e-06 & 2.804e-18 & 2.827e-10 & 3.043e-07 & 2.273e-10 & 1.296e-06 & 5.100e-05 & 1.355e-19 & 5.894e-05 & 1.300e-06 \\
                ~ & ~ & 4.706e-10 & 6.621e-11 & 4.249e-10 & 4.016e-10 & 1.946e-05 & 8.091e-05 & 8.342e-09 & 4.460e-12 & 1.885e-09 & 2.745e-10 & 1.143e-09 & 3.393e-10 \\
                \midrule
			\multirow{4}{*}{RCB} & \multirow{2}{*}{mean} & 1.633e+02 $+$ & 1.666e+02 $+$ & 1.617e+02 $+$ & 1.666e+02 $+$ & 1.600e+02 $+$ & 1.684e+02 $+$ & 1.668e+02 $+$ & 1.599e+02 $+$ & 1.683e+02 $+$ & 1.670e+02 $+$ & 1.713e+02 $+$ & 1.658e+02 $+$ \\
                ~ & ~ & \textbf{1.594e+02} $-$ & 1.671e+02 $+$ & 1.610e+02 $+$ & 1.673e+02 $+$ & 1.687e+02 $+$ & 1.720e+02 $+$ & 1.610e+02 $+$ & 1.654e+02 $+$ & 1.595e+02 $\approx$ & 1.596e+02 $\approx$ & 1.595e+02 $\approx$ & 1.596e+02 \\
                ~ & \multirow{2}{*}{std} & 1.289e+00 & 6.055e-01 & 2.090e+00 & 6.055e-01 & 6.105e-01 & 2.189e+00 & 1.293e+00 & 4.878e-01 & 2.654e+00 & 1.295e+00 & 4.189e+00 & 2.503e+00 \\
                ~ & ~ & 6.632e-02 & 8.134e-01 & 1.503e+00 & 1.830e+00 & 2.097e+00 & 6.208e+00 & 1.725e+00 & 7.146e-01 & 1.233e-01 & 3.588e-01 & 1.218e-01 & 2.726e-01 \\
                \midrule
			\multirow{4}{*}{SRD} & \multirow{2}{*}{mean} & 2.991e+03 $-$ & 2.988e+03 $-$ & 3.138e+03 $+$ & \textbf{2.988e+03} $-$ & 3.006e+03 $-$ & 3.095e+03 $+$ & 3.051e+03 $-$ & 3.196e+03 $+$ & 4.152e+03 $+$ & 3.445e+03 $+$ & 3.078e+06 $+$ & 2.916e+04 $+$ \\
                ~ & ~ & 3.072e+03 $+$ & 3.076e+03 $+$ & 3.024e+03 $-$ & 3.049e+03 $-$ & 3.435e+03 $+$ & 2.658e+06 $+$ & 3.150e+03 $+$ & 3.013e+03 $\approx$ & 3.084e+03 $+$ & 3.022e+03 $+$ & 3.060e+03 $+$ & 3.013e+03 \\
                ~ & \multirow{2}{*}{std} & 2.387e+00 & 4.226e-01 & 5.258e+01 & 2.239e-01 & 4.417e+00 & 3.361e+01 & 4.108e+01 & 6.322e+01 & 9.313e+02 & 5.593e+02 & 2.002e+06 & 1.314e+05 \\
                ~ & ~ & 1.347e+01 & 1.858e+01 & 1.007e+01 & 5.891e+01 & 3.358e+02 & 1.999e+06 & 4.632e+01 & 1.213e+01 & 2.022e+01 & 5.815e+00 & 1.786e+01 & 6.966e+00 \\
                \midrule
                \multicolumn{2}{c}{\multirow{2}{*}{$+$/$\approx$/$-$:}} & 5/0/1 & 4/1/1 & 6/0/0 & 3/0/3 & 2/3/1 & 6/0/0 & 4/1/1 & 6/0/0 & 5/0/1 & 3/1/2 & 6/0/0 & 6/0/0 \\ 
                ~ & ~ & 5/0/1 & 5/0/1 & 3/2/1 & 3/1/2 & 6/0/0 & 6/0/0 & 5/0/1 & 3/1/2 & 4/2/0 & 2/3/1 & 4/2/0 & - \\
            \midrule
                \multicolumn{2}{c}{\multirow{2}{*}{Ave ranks:}} & 13.3 & 10.5 & 17.0 & 7.5 & 5.3 & 20.1 & 12.8 & 12.0 & 16.8 & 9.1 & 22.0 & 15.0 \\
                ~ & ~ & 12.0 & 11.6 & 9.1 & 13.5 & 21.3 & 21.8 & 11.6 & 8.8 & 9.8 & 5.3 & 8.0 & \textbf{5.1} \\
		  \bottomrule
		\end{tabular}
	}
\end{sidewaystable}

\section{Discussion} \label{sec:5}
This section commences with a theoretical analysis of the computational complexity of ZSO, then, the performance analysis on three benchmarks is summarized, and finally, we project the future trajectory of the EC techniques in the context of the LLM era.

\subsection{Computational complexity analysis} \label{sec:5.1}
Assuming the population size is $N$, the maximum iteration is $T$, and the dimension size is $D$. The computational complexity of the three main operators is delineated as follows:
\begin{itemize}
  \item Swarm initialization: $O(N\times D)$
  \item Prey-predator interaction operator for the current population: $O(N\times D)$
  \item Social flocking operator for the current population: $O(N\times D + N\times D)$:=$O(N\times D)$
\end{itemize}
Therefore, the overall computational complexity of ZSO can be expressed as $O(N\times D)$ + $O(T\times N\times D)$:=$O(T\times N\times D)$ totally. Compared with the popular MAs listed in Table \ref{tbl:5.1.1}, since the fitness-based sort operator is not involved in the proposed ZSO, it has a lighter computational burden than the GWO and is equal to other optimizers.
\begin{table}[htbp]
	\scriptsize
	\centering
	\renewcommand\arraystretch{1.4}
	\caption{Popular MAs and corresponding computational complexity.}
	\label{tbl:5.1.1}
		\begin{tabular}{ccc}
            \toprule
                Alg. & Computational complexity \\
            \midrule
                DE \cite{Storn:97} & $O(T\times N\times D)$ \\
                PSO \cite{Kennedy:95} & $O(T\times N\times D)$ \\
                GWO \cite{Seyedali:14} & $O(T\times N \times log(N) \times D)$ \\
                WOA \cite{Andrew:16} & $O(T\times N \times D)$ \\
                GJO \cite{Nitish:22} & $O(T\times N \times D)$ \\
		  \bottomrule
		\end{tabular}
\end{table}

\subsection{Performance analysis on CEC2014 benchmark functions} \label{sec:5.2}
The benchmark functions in the CEC2014 exhibit various characteristics, including unimodal ($f_1$-$f_3$), multimodal ($f_4$-$f_{16}$), hybrid ($f_{17}$-$f_{22}$), and composite ($f_{23}$-$f_{30}$) functions. This diversity ensures that the optimization challenges presented within this suite comprehensively reflect the performance capabilities of the investigated algorithms. We can observe the convergence status of all MAs on CEC2014 benchmark functions in Figures \ref{fig:4.2.1.1} and \ref{fig:4.2.1.2}, while the statistical results are summarized in Tables \ref{tbl:4.2.1.1} and \ref{tbl:4.2.1.2}. From these experimental results, the LLM-assisted ZSO is highly competitive with other compared MAs, particularly in unimodal, multimodal, and hybrid functions. In most instances, ZSO and its variants perform better than the compared MAs in most cases, which demonstrates the superiority of the proposed ZSO. Additionally, the summary of the statistical analysis and the computation of the average ranks further corroborate the efficiency and robustness of ZSO-derived algorithms.

However, the observed under-performance of the ZSO in composite functions, when compared with other algorithms like HHO, AO, ArchOA, AOA, GJO, FOX, CDO, and COA is a significant aspect that warrants attention. ZSO\_Gauss is significantly worse than the mentioned MAs in composite functions. We infer that the architecture of ZSO and the search operators designed by the LLM may not be suitable for dealing with composite functions. To contextualize this phenomenon, the No Free Lunch Theorem (NFLT) offers a pertinent explanation. NFLT states that if an algorithm excels in solving a particular class of problems, it inherently comes at the cost of reduced performance in other problem classes. This theoretical perspective suggests that, on average, any pair of algorithms will exhibit similar performance across all observable problem domains.

\subsection{Performance analysis on CEC2022 benchmark functions} \label{sec:5.3}
As one of the latest benchmark suites, the CEC2022 benchmark contains twelve challenging optimization functions for evaluating the MAs thoroughly. The optimization in this benchmark suite allows performance evaluation in complex environments. Similarly, ZSO and its variants perform competitiveness in unimodal, basic, and hybrid functions. This competitiveness is further highlighted by the rapid convergence speed observed in Figure \ref{fig:4.2.2.1}, which presents the excellent exploitation ability of ZSO-derived algorithms. The final optimum found by ZSO\_Gauss also exhibits a dominant performance over the compared algorithms, which is owing to a clever balance between exploitation and exploration. 

\subsection{Performance analysis on engineering problems} \label{sec:5.4}
Through the convergence curves presented in Figure \ref{fig:4.2.3.1} and the statistical analysis summarized in Table \ref{tbl:4.2.3.1}, the proposed ZSO has outstanding performance in real-world simulation problems. We infer that the ZSO-derived algorithms may have extensively robust performance in real-world optimization tasks. 

\subsection{Prospects of the metaheuristics community under LLM era} \label{sec:5.5}
In this paper, we propose a novel MA ZSO through the LLM ChatGPT-3.5 and the CRISPE framework. With slight human-interacted modification, the proposed ZSO and its variants have competitive performance with state-of-the-art MAs. Thus, using LLM to design novel metaphor-based MAs is hopeful. Moreover, during the LLM era, what benefits can LLM bring to the metaheuristic community, and what advantages can metaheuristics contribute to the LLM? With these questions, we ask ChatGPT-3.5 again with the CRISPE-based prompt: \textit{Act as an expert in the fields of MAs and large language models. We want to summarize the potential benefits of introducing the large language model to the metaheuristic community and what advantages can MAs contribute to the large language model. Provide some potential prospects.}

According to the ChatGPT-3.5, the responses are as follows: Integrating large language models (LLMs) into the realm of MAs holds significant promise and can lead to several mutually beneficial outcomes. Let's explore the potential benefits and advantages of introducing LLMs to the metaheuristic community and vice versa:

\textbf{Benefits of Introducing LLMs to Metaheuristics}:
\begin{itemize}
  \item \textbf{Enhanced Problem Representation}: LLMs can improve the representation of problem instances, offering a more nuanced and context-aware understanding of the optimization landscape. Natural Language Processing (NLP) capabilities in LLMs allow for a more expressive and human-like problem description, aiding in problem definition and interpretation.
  \item \textbf{Knowledge Integration}: Pre-trained LLMs can incorporate vast amounts of domain-specific knowledge, providing metaheuristics with a comprehensive understanding of the problem space. Fine-tuning LLMs on specific problem instances can lead to the extraction of relevant features and patterns, aiding metaheuristics in decision-making.
  \item \textbf{Dynamic Adaptation}: LLMs can facilitate adaptive metaheuristics by dynamically adjusting their parameters based on real-time information and evolving problem conditions. Continuous learning capabilities of LLMs can enhance adaptability, allowing metaheuristics to respond effectively to changes in the optimization landscape.
  \item \textbf{Multi-Objective Optimization}: LLMs can assist in handling multi-objective optimization problems by comprehensively understanding and representing the trade-offs among conflicting objectives. The natural language understanding in LLMs can help in formulating and communicating complex objective functions in a more human-readable manner.
\end{itemize}

\textbf{Advantages of MAs for LLMs}:
\begin{itemize}
  \item \textbf{Training Data Augmentation}: MAs can be employed to generate diverse and challenging training instances for LLMs, improving their robustness and generalization capabilities. Exploration-exploitation strategies in metaheuristics can contribute to more effective data sampling during training.
  \item \textbf{Hyperparameter Tuning}: Metaheuristics can be utilized to optimize the hyperparameters of large language models, enhancing their performance on specific tasks. Evolutionary algorithms and swarm intelligence can efficiently search the hyperparameter space for optimal configurations.
  \item \textbf{Transfer Learning and Adaptation}: Metaheuristics can aid in the transfer learning process for LLMs by facilitating the adaptation of pre-trained models to new tasks or domains. Adaptive algorithms can optimize the fine-tuning process, ensuring faster convergence and improved task-specific performance.
  \item \textbf{Resource Allocation and Scaling}: Metaheuristics can optimize the allocation of computational resources for training large language models, improving efficiency and reducing training time. Scalability algorithms can ensure that LLMs can handle increased model sizes and datasets effectively.
\end{itemize}

In summary, the collaboration between LLMs and metaheuristics holds immense potential for advancing the capabilities of optimization techniques and creating more intelligent, adaptive, and user-friendly solutions across a wide range of applications. This interdisciplinary synergy can pave the way for groundbreaking developments in the field of artificial intelligence and optimization.

\section{Conclusion} \label{sec:6}
In this paper, we propose a groundbreaking animal-inspired MA named ZSO which is founded upon the principles of ChatGPT-3.5 and the CRISPE framework tailored for prompt design. Depending on the basic framework of ZSO, the human-interacted design is embedded into the original ZSO to contribute to the development of ZSO's variants. Through comprehensive numerical experiments conducted on CEC2014 benchmark functions, CEC2022 benchmark functions, as well as engineering optimization, the effectiveness and robustness of ZSO are confirmed. 

In summary, leveraging LLM to generate a novel MA with the CRISPE framework is feasible and efficient. As the paper concludes, we present several prospective avenues for future research, aiming to foster the growth and healthy development of the metaheuristics community in the era of LLMs.

\section{Statement and Declarations}

\subsection{Competing interest}
The authors declare that they have no known competing financial interests or personal relationships that could have appeared to influence the work reported in this paper.

\subsection{Data availability}
The source code of this research can be downloaded from \url{https://github.com/RuiZhong961230/ZSO}.

\subsection{Acknowledgement}
This work was supported by JST SPRING Grant Number JPMJSP2119.

\bibliographystyle{elsarticle-num}
\bibliography{paper}
\end{document}